\documentclass{article}

    \PassOptionsToPackage{numbers, compress}{natbib}

\usepackage[final]{neurips_2024}

\usepackage[utf8]{inputenc} 
\usepackage[T1]{fontenc}    
\usepackage{hyperref}       
\usepackage{url}            
\usepackage{graphicx}
\usepackage{booktabs}       
\usepackage{amsfonts}       
\usepackage{nicefrac}       
\usepackage{microtype}      
\usepackage{amsmath}
\usepackage{subfig}
\usepackage{multirow}
\usepackage{mathabx}
\usepackage{adjustbox}
\usepackage{graphicx}
\usepackage{rotating}
\usepackage[table]{xcolor}

\usepackage{amsmath,amsfonts,bm}
\usepackage{bbm}
\usepackage[mathscr]{eucal}

\def\1{\bm{1}}

\def\<{\langle}
\def\>{\rangle}

\def\mD{{\bm{D}}}
\def\mE{{\bm{E}}}
\def\mF{{\bm{F}}}

\def\mI{{\bm{I}}}

\def\mL{{\bm{L}}}

\def\mS{{\bm{S}}}

\DeclareMathAlphabet{\mathsfit}{\encodingdefault}{\sfdefault}{m}{sl}
\SetMathAlphabet{\mathsfit}{bold}{\encodingdefault}{\sfdefault}{bx}{n}

\let\hat\relax
\newcommand{\hat}{\widehat}

\title{Latent Intrinsics Emerge from Training to Relight}

\author{Xiao Zhang$^{1}$ \textbf{\enspace\enspace William Gao$^{1}$ \enspace\enspace Seemandhar Jain$^{2}$ \enspace\enspace Michael Maire$^{1}$}\\\textbf{\enspace\enspace D.A. Forsyth$^{2}$ \enspace\enspace Anand Bhattad$^{3}$}
\vspace{1mm}\\$^1$University of Chicago \enspace\enspace $^2$ University of Illinois Urbana Champaign \\$^3$Toyota Technological Institute at Chicago\vspace{1mm}\\\texttt{zhang7@uchicago.edu \enspace\enspace   bhattad@ttic.edu\vspace{1mm}}
\\\texttt{\href{https://sites.google.com/uchicago.edu/latent-intrinsic/home}{https://sites.google.com/uchicago.edu/latent-intrinsic/home}}
}

\begin{document}

\maketitle
\begin{figure}[ht]
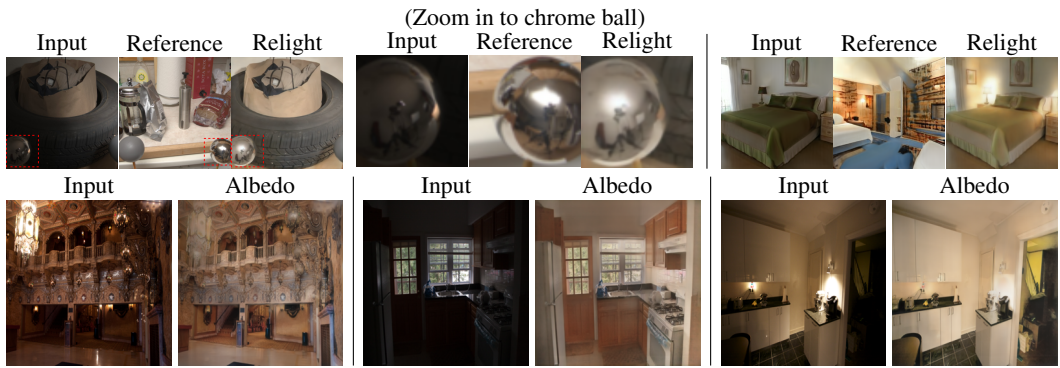

    \centering
    \input{figure/teaser_relight_light}
    \par\vspace{2pt}
    \input{figure/teaser_albedo_light}
    \caption{We describe a purely data-driven image relighting model.  Our model recovers latent variables representing
      scene intrinsic properties from one image, latent variables representing lighting from another, then applies the lighting
      to the intrinsics to produce a relighted scene ({\bf top row}).  There is no physical model of intrinsics, extrinsics or their interaction.  Our model relights images of real scenes with SOTA accuracy and is more accurate
      than current supervised methods.  Note how, for the chrome ball detail in {\bf top center}, the specular reflections on the chrome ball (which give an approximate environment map) change when the extrinsics are changed.  Note how our model ascribes lighting to visible luminaires when it can ({\bf top right}), despite the absence of any physical model.    A physical model accounts only for effects in that model, and most physical models of surfaces are approximate; in contrast, a latent intrinsic model accounts for whatever produces substantial effects in training data.  Latent intrinsics yield albedo
      in a natural fashion (light the scene with an appropriate illuminant). {\bf Bottom row} shows  SOTA albedo estimates recovered from our latent intrinsics.
    }
    \label{fig:teaser}
    \vspace{-5pt}
\end{figure}

\begin{abstract}
  Image relighting is the task of showing what a scene from a source image would look like if illuminated differently.  Inverse graphics schemes recover an explicit representation of geometry and a set of chosen intrinsics, then relight with some form of renderer.  However error control for inverse graphics is difficult, and inverse graphics methods can represent only the effects of the chosen intrinsics. This paper describes a relighting method that is entirely data-driven, where intrinsics and lighting are each represented as latent variables.  Our approach produces SOTA relightings of real scenes, as measured by standard metrics.  We show that albedo can be recovered from our latent intrinsics without using any example albedos, and that the albedos recovered are competitive with SOTA methods.
\end{abstract}

\section{Introduction}
\label{sec:intro}
Relighting -- taking an image of a scene, then adjusting it so it
looks as though it had been taken under another light -- has a range
of applications, including commercial art (e.g., photo enhancement) and data augmentation (e.g., making vision models robust to varying illumination). 
As a technical problem, relighting is very hard indeed, likely because
how a scene changes in appearance when the light is changed can depend
on complex surface details (grooves in screws; bark on trees; wood
grain) that are hard to capture either in geometric or surface models.

One common approach to relighting a scene is to infer scene characteristics (geometry, surface properties) using inverse graphics methods, then render the scene with a new light source. This approach is fraught with difficulties, including the challenge of selecting which material properties to infer and managing error propagation. These methods perform best in outdoor scenes with significant shadow movements but struggle with indoor scenes where interreflections create complex effects (Section~\ref{sec:relight}).

As this paper demonstrates, a purely data-driven method offers an attractive alternative. A source scene, represented by an image, is encoded to produce a latent representation of intrinsic scene properties. A source illumination, represented by another image, is encoded to produce a latent representation of illumination properties. These intrinsic and extrinsic properties are combined and then decoded to produce the relighted image. As a byproduct of this training, we find that the latent representation of intrinsic scene properties behaves like an albedo, while another latent representation acts as a lighting controller.

Our model can capture complex scene characteristics without explicit supervision by capturing intrinsic properties as latent phenomena, making it particularly appealing.  In contrast to a physical model, we are not required to choose which effects to capture.  This latent approach reduces the need for detailed geometric and surface models, simplifies the learning process, and enhances the model's ability to generalize to diverse and unseen scenes. This makes it highly applicable to a wide range of real-world scenarios.

{\bf Contributions:} We present the first fully data-driven relighting method applicable to images of real complex scenes. Our approach requires no explicit lighting supervision, learning to relight using paired images alone. We demonstrate that this method effectively trains and generalizes, producing highly accurate relightings. Furthermore, we demonstrate that albedo-like maps can be generated from the model without supervision or prior knowledge of albedo-like images. These intrinsic properties emerge naturally within the model. We validate our model on a held-out dataset, applying target lighting conditions from various scenes to assess its generalization capability and precision in real-world scenarios (Section~\ref{sec:relight}).

\section{Related Work}
\label{sec:related}
\noindent{\bf Intrinsic Images.} Humans have been known to perceive scene
properties independent of lighting since at least
1867~\cite{Helmholtz,Hering,beckbook,Gilchristbook}.
In computer vision, the idea dates to Barrow and
Tenenbaum~\cite{barrowtenenbaum} and comprises at least depth, normal,
albedo, and surface material maps.   Depth and normal estimation are now
well established (eg~\cite{kar20223d}).  There is a rich literature on
albedo estimation (dating to 1959~\cite{nn8146,nn8147}!).
A detailed review appears in~\cite{forsyth2021intrinsic}, which breaks out methods as to
what kinds of training data they see. Early methods do not see any form of training data, but more recently both
CGI data and manual annotations of relative lightness (labels) have become available. Early efforts, such as SIRFS~\citep{barron2014shape}, focused on using shading information to recover shape, illumination, and reflectance, highlighting the importance of modeling these factors for intrinsic image analysis. 
Recent strategies include: deep networks trained on synthetic data~\cite{li2018cgintrinsics, janner2017self, fan2018revisiting}; and
conditional generative models~\cite{kocsis2023intrinsic}. 

The weighted human
disagreement ratio (WHDR) evaluation framework was introduced by
\cite{bell14intrinsic_short} using the IIW dataset.  This is a dataset of human judgments that compare the absolute lightness at pairs of points in real images.  Each pair is labeled with one of three cases (first lighter; second lighter; indistinguishable)
and a weight, which captures the certainty of labelers.  
One evaluates by computing a weighted comparison of algorithm predictions with
human predictions; 
WHDR scores can be improved by postprocessing because most methods produce albedo fields with very slow gradients,
rather than piecewise constant albedos.  \cite{bi20151} demonstrate the value of ``flattening'' albedo (see also
\cite{nestmeyer2017reflectance}); \cite{EGSR18} employ a fast bilateral filter~\cite{bilat} to obtain significant
improvements in WHDR.  

\noindent{\bf Using Intrinsic Images for Relighting.}
Bhattad and Forsyth~\cite{bhattad2022cut}
demonstrated that intrinsic images could be used for reshading
inserted objects. This approach can be extended by adjusting the
shading in both the foreground and background to eliminate
discrepancies~\cite{careaga2023intrinsic}. Intrinsic images and
geometry-aware networks have been used for multi-view
relighting~\cite{philip2019multi}. StyLitGAN~\cite{StyLitGAN} introduced a method to
relight images by identifying directional vectors in the latent space
of StyleGAN, but can only relight StyleGAN generated images and requires explicit albedo and shading to guide relighting.  It can be extended to real images using a GAN inversion, but does not generalize~\cite{bhattad2023make}. 
LightIt~\cite{kocsis2024lightit} controls lighting changes in image generation using diffusion models, by conditioning on shading and normal maps to achieve consistent and controllable lighting.
Like these methods, we use
intrinsics and extrinsics to relight, but ours are latent, with no explicit physical meaning.

\noindent{\bf Color Constancy.} Image color is ambiguous: a green pixel
could be the result of a white light on a green surface, or a green light
on a white surface.  Humans are unaffected by this
ambiguity (eg \cite{Hering,beckbook}; recent review
in~\cite{ccreview}). There is extensive computer
vision literature; a recent review appears in~\cite{pamicc}.  We
do not estimate illumination color but estimate a single color
correction (Section~\ref{sec:relight}).

\noindent{\bf Lighting Estimation and Representation.}
Accurate lighting representation is crucial for tasks like object
insertion and relighting. Traditional methods used parametric models
such as environment maps and spherical harmonics to represent
illumination~\cite{debevec1998RSO,
  ramamoorthi2001relationship}. Debevec's seminal
work~\cite{debevec1998RSO} on recovering environment maps from
images of mirrored spheres set the foundation for many subsequent
works. Methods by Karsch et al.~\cite{karsch2011rendering,
  karsch2014automatic}, Gardner et al.~\cite{gardner2017learning,
  gardner2019deep}, Garon et al.~\cite{garon2019fast} and Weber at al.~\cite{weber2022editable}  advanced the field by using learned models to
recover parametric, semi-parametric or panoramic representations of illumination. Recent approaches
include representing illumination fields as dense 2D grids of
spherical harmonic sources~\cite{li2020inverse, li2022physically} or
learning 3D volumes of spherical
Gaussians~\cite{wang2021learning}. These methods can model complex
light-dependent effects but require extensive CGI datasets for
training~\cite{roberts2021hypersim, li2021openrooms}. Our approach
diverges by not relying on labeled illumination representations or CGI data, instead producing
abstract representations of illumination through deep features without
specific physical interpretations. 

\noindent{\bf Image-based Relighting.} Other works focus on portrait relighting
using deep learning~\cite{sun2019single, zhou2019deep,
  nestmeyerlearning, sengupta2021light}, which are typically
specialized to faces and trained on paired or light-stage
data. Self-supervised methods for outdoor image relighting leverage
single-image decomposition with parametric outdoor illumination,
benefiting from simpler lighting conditions dominated by sky and
sunlight~\cite{yu2020self, liu2020learning}. ~\citep{hu2020sa} introduced a self-attention autoencoder model to re-render a source image to match the illumination of a guide image, focusing on separating scene representation and lighting estimation with a self-attention mechanism for targeted relighting. Similarly, ~\citep{yang2021s3net} proposed a depth-guided image relighting, which combines source and guide images along with their depth maps to generate relit images.  In contrast, our work
shows that intrinsic properties relevant to relighting can emerge
naturally from training to relight, facilitating complex scene
relighting without the need for explicit lighting estimation. We compare with both ~\citep{hu2020sa} and ~\citep{yang2021s3net} for relighting capabilities on real scenes.

\noindent{\bf Emergent Intrinsic Properties.}
Bhattad et al.~\cite{bhattad2024stylegan} and Du et al.~\cite{du2023generative}
demonstrate that intrinsic images can be extracted from generative models using a small intrinsic image dataset obtained from pretrained off-the-shelf intrinsic image models. Our work explores how intrinsic image properties emerge as a result of training a model for relighting, without the need for an intrinsic image dataset.

\section{Learning Latent Intrinsic from Relighting.}

\begin{figure}[t]
\captionsetup{singlelinecheck=false}
    \centering
\includegraphics[width=1.0\textwidth]{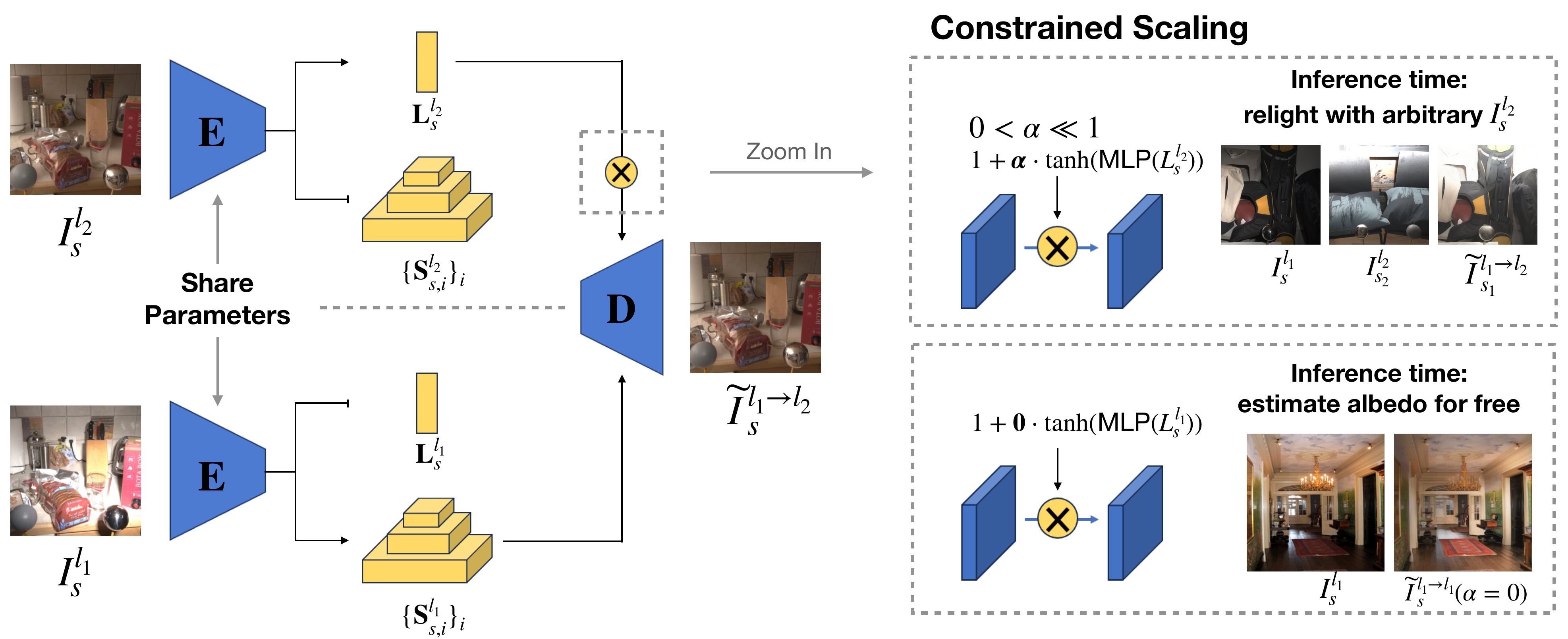}
    \caption{The network diagram of our relighting model. The model functions as an autoencoder, comprising an encoder $\mE$ and a decoder $\mD$. \textbf{\textit{Left Half}}: The encoder $\mE$ maps input image $\mI_s^l$, captured under scene $s$ and lighting $l$, to low-dimensional extrinsic features $\mL_s^l$ and set of intrinsic features map $\{S_{s,i}^l\}_i$. The decoder $\mD$ then generates new images based on these intrinsic and extrinsic representations. \textbf{\textit{Right Half}}: We employ \textit{constrained scaling} for the injection of  $\mL_{s}^l$, utilizing $0<\alpha\ll 1$ to regularize the information passed from $\mL_s^l$, thereby enforcing a low-dimensional parameterization of the extrinsic features. We train our system to relight target images given input paired with images captured under the same scene $s$. During inference, our model demonstrates the ability to generalize to arbitrary reference images for relighting and can estimate albedo for free.}
    \label{fig:enter-label}
    \vspace{-8pt}
\end{figure}

Our relighting model can be seen as a form of autoencoder.  One
encoder computes a latent representation of scene intrinsics from an
image of a target scene; another computes a latent representation of scene
extrinsics from an image of a placeholder scene in the reference lighting.  These are combined,
then decoded into a final image of the target scene in the reference
lighting.  Losses impose the requirements that (a) the final image is
right and (b) the latent intrinsics computed for a scene are not
affected by illumination.   The procedure for combining intrinsics and
extrinsics is carefully designed to make it very difficult for
intrinsic features of the placeholder scene to ``leak'' into the final image.

\subsection{Model structure}

  {\bf Decoder setup:} Write  $\mI_{s}^{l} \in \mathbf{R}^{H\times  W\times 3}$ for the input image,
captured from scene $s$ with lighting configuration $l$.  Training
uses pairs $\mI_{s}^{l_1}$ and $\mI_{s}^{l_2}$, representing the same
scene $s$ under different lighting conditions $l_1$ and $l_2$.   The
model {\em does not see} detailed lighting information (for example,
the index of the lighting) during training, because
standardizing lighting settings across various scenes is often
impractical. 

Write $E$ for the encoder,  $D$ for the decoder. The encoder must
produce the intrinsic and extrinsic representations from the input
image.  Write ${\mS^{l}_{s,i} \in \mathbf{R}^{(H_i\times W_i)\times C_i} }$
for spatial feature maps yielding the intrinsic representation, with
$i$ for the layer index, and $\mL^l_s \in \mathbf{R}^C$ for extrinsic
features; we have:
\begin{eqnarray}E(\mI_{s}^l) := \{\mS^{l}_{s,i}\}_i, \mL^l_s\end{eqnarray}
We apply L2 normalization along the feature channel to both
sets of features.  During training, we add random Gaussian noise
to the input image to enhance semantic scene understanding
capabilities: 
\begin{eqnarray}E(\mI_{s}^l + \sigma\epsilon) := \{\mS^{l}_{s,i}\}_i, \mL^l_s\end{eqnarray}

{\bf Decoder setup:} The decoder $D$ relights $\mI_{s}^{l_1}$ using extrinsic features extracted from $\mI_{s}^{l_2}$:
\begin{eqnarray}D( \{\mS^{l_1}_s\}, \mL^{l_2}_s) := \Tilde{\mI}_s^{l_1\rightarrow l_2}\end{eqnarray}
We optimize the autoencoder using a pixel-wise loss on both relighted and reconstructed images:
\begin{eqnarray}
\mathcal{L}_{\text{relight}} := \mathcal{L}_{\text{pixel}}(\Tilde{\mI}_s^{l_1\rightarrow l_2}, \mI_s^{l_2}) + \mathcal{L}_{\text{pixel}}(\Tilde{\mI}_s^{l_2\rightarrow l_2}, \mI_s^{l_2})
\label{eqn:relight}
\end{eqnarray}
where $\mathcal{L}_{\text{pixel}}$ represents the pixel-wise losses:
L2 distance on pixels; structural similarity index (SSIM)
~\citep{wang2004image}; and l2 distance on image spatial gradient
(weights 10, 0.1 and 1 respectively). 

\subsection{Intrinsicness}

{\bf Intrinsicness:}  Our model should report the same latent
intrinsic for the same scene in different lightings, so we apply the
following loss to the encoder:
\begin{eqnarray}
   \mathcal{L}_{\text{intrinsic}} := \sum_{i}\|\mS_{s,i}^{l_1} - 
   \mS_{s,i}^{l_2}\|_2 + \text{1e-3}\cdot \mathcal{L}_{\text{reg}}(\mS_{s,i}^{l_1})
   \label{eqn:intrinsic_loss}
\end{eqnarray}
where $\mathcal{L}_{\text{reg}}$ is a regularization term on intrinsic features, defined as follows:
\begin{eqnarray}
   \mathcal{L}_{\text{reg}}(\mS) &:=& 
   \|R(\mS) - R(\hat{\mS})\|_2 \\ 
   R(\mS) &:=& 
   \log\det\left(\mI + \frac{d}{n\lambda^2}\mS^{\top}\mS\right)
    \label{eqn:reg}
\end{eqnarray}
where $R(\mS)$ is the coding rate~\citep{yu2020learning} for a matrix
$\mS \in \mathbb{R}^{n\times d}$ with each row l2 normalized, under a
distortion constant $\lambda$.  $\hat{\mS}$ is a random matrix with
the same shape of $\mS$ and each row of $\hat{\mS}$ is sampled from
uniform hyperspherical distribution at the start of learning. In Eqn.\ref{eqn:intrinsic_loss},
$R(\hat{\mS})$ serves as the optimization target of $R(\mS)$ to
encourage the $\mS$ to uniformly spread out in the hyperspherical
space.  This strategy is now widely used in self-supervised learning;
without the regularization term, the model can minimize the feature
distance by simply collapsing the distribution of $\mS_{s,i}^l$ with
small variance, which will not yield effective lighting invariance.  

\subsection{Combining intrinsics and extrinsics}

The placeholder scene is necessary to communicate illumination to the
model, but has important nuisance features.  Intrinsic information
from this scene could ``leak'' into the final image, spoiling
results.    We introduce \textit{constrained scaling}, 
a structural bottleneck that restricts the amount of information
transmitted from the learned extrinsic features.

Write $\mF \in \mathbf{R}^{h \times w \times c}$ for the feature map
fed to the decoder. Constrained scaling combines intrinsic and
extrinsic features by
\begin{eqnarray}
   \Tilde{\mF} := \mF \odot \left(1 + \alpha\cdot \text{tanh}\left({\text{MLP}}\left(\mL^l_s\right)\right)\right)
   \label{eqn:constraint_scale}
\end{eqnarray}
where MLP, a series of fully connected layers with non-linear
activation, aligns $\mL_s^l$ to the latent channel dimension of $\mF$
and $\alpha\ll 1$ is a small non-negative scalar (we use 5e-3). This
approach means that any single extrinsic feature vector has little
effect on the feature -- for an effect, the extrinsics must be pooled
over multiple locations.  Illumination fields tend to be spatially
smooth, supporting the insight that enforced pooling is a good idea.

Constrained scaling compresses  latent vectors into a very small
numerical range, making learning difficult.
We use a regularizer to promote a uniform distribution of $\mL_s^l$,
which improves optimization. In particular, we have
\begin{eqnarray}
   \mathcal{L}_{\text{extrinsic}} := \mathcal{L}_{\text{reg}}(\mL_{s}^{l})
   \label{eqn:extrinsic_loss}
\end{eqnarray}
By choosing $\alpha \ll 1$  and training model with uniform
regularization term Eqn.\ref{eqn:extrinsic_loss}, we effectively push
the lighting code to uniformly spread over [$-\alpha$, $\alpha$] where
the absolute value of each channel indicates the strength of the
light. As a side effect, by setting $\alpha = 0$ to disable the
contribution of the lighting code, we get image albedo estimation from
our model for free. 

Our final training objective is weighted combination of all individual loss terms:
\begin{eqnarray}
   \mathcal{L}:= \mathcal{L}_{\text{relight}} + \text{1e-1}\cdot \mathcal{L}_{\text{intrinsic}} + \text{1e-4}\cdot \mathcal{L}_{\text{extrinsic}}
\end{eqnarray}

\section{Experiments}
We will first provide a brief description of our experimental procedure (Sec~\ref{sec:details}), followed by a discussion on how we evaluate the various relighting capabilities of our approach, including its strong generalization across datasets with different distributions (Sec~\ref{sec:relight}). Finally, we will present the emergent albedo that is recovered from the latent intrinsic without using any albedo-like images (Sec~\ref{sec:albedo_eval}).

\subsection{Experiment Details}
\label{sec:details}
\textbf{Training Details} 
We train our model using the MIT
multi-illumination dataset~\citep{murmann2019dataset}, which includes
images of 1,015 indoor scenes captured under 25 fixed lighting, totaling 25,375 images. We follow the official data split
and train our model on the 985 training scenes. During training, we
randomly sample pairs of images from the same scene under different lighting conditions and
perform random spatial cropping, with the crop ratio randomly selected
between 0.2 and 1.0, followed by resizing the cropped image to a
resolution of 256x256. For further details, please refer to our
appendix.  

\subsection{Evaluating image relighting}
\label{sec:relight}
\begin{figure}[t]
    \centering
    \begin{minipage}{0.07\linewidth}
        \centering
        \textbf{Input}
        \textcolor{black}{\rule[1ex]{\linewidth}{2pt}}
        \includegraphics[width=1.0\textwidth]{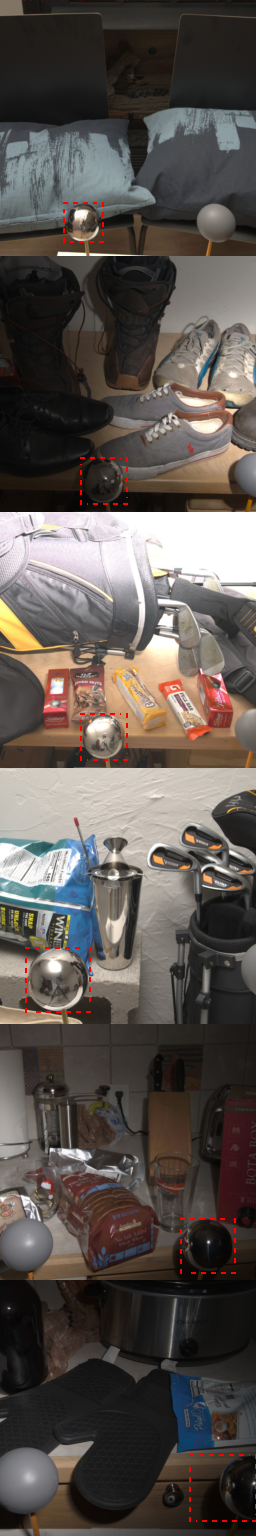}
    \end{minipage}
    \begin{minipage}{0.07\linewidth}
        \centering
        \textbf{Ref}
        \textcolor{black}{\rule[1ex]{\linewidth}{2pt}}
        \includegraphics[width=1.0\textwidth]{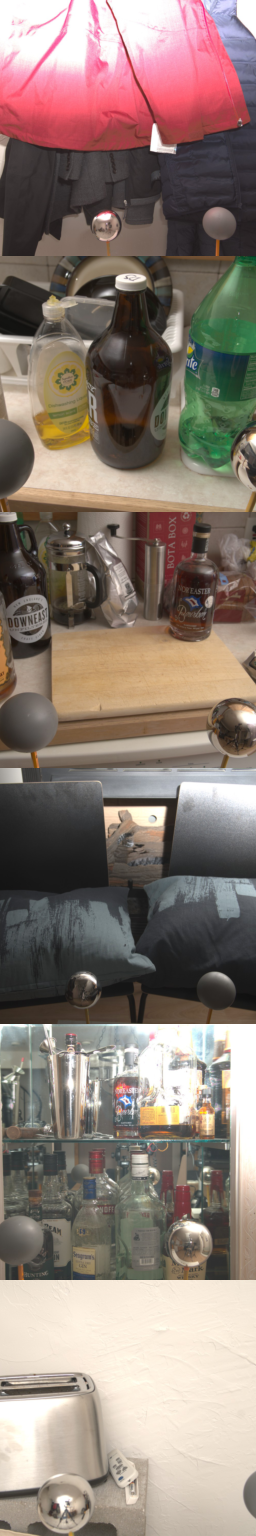}
    \end{minipage}
    \begin{minipage}{0.07\linewidth}
        \vspace{-10pt}
        \centering
        {\small \textbf{SAAE sup}}
        \textcolor{black}{\rule[1ex]{\linewidth}{2pt}}
        \includegraphics[width=1.0\textwidth]{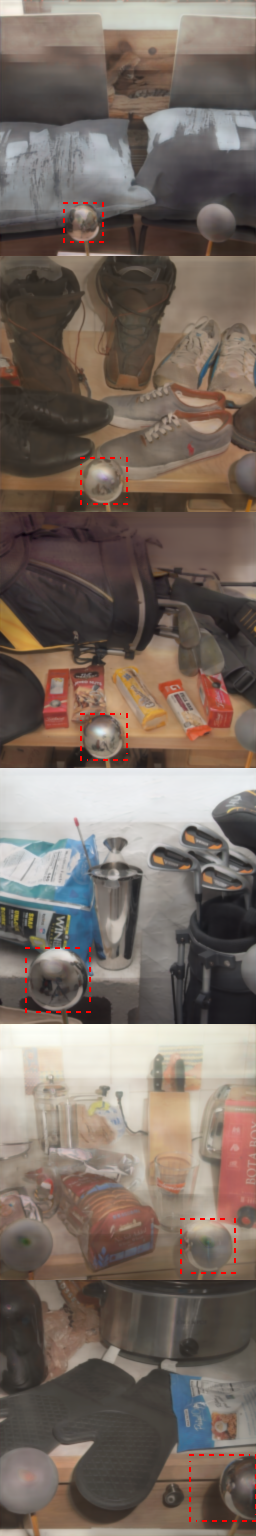}
    \end{minipage}
    \begin{minipage}{0.07\linewidth}
        \vspace{-10pt}
        \centering
        {\small \textbf{SAAE unsup}}
        \textcolor{black}{\rule[1ex]{\linewidth}{2pt}}
        \includegraphics[width=1.0\textwidth]{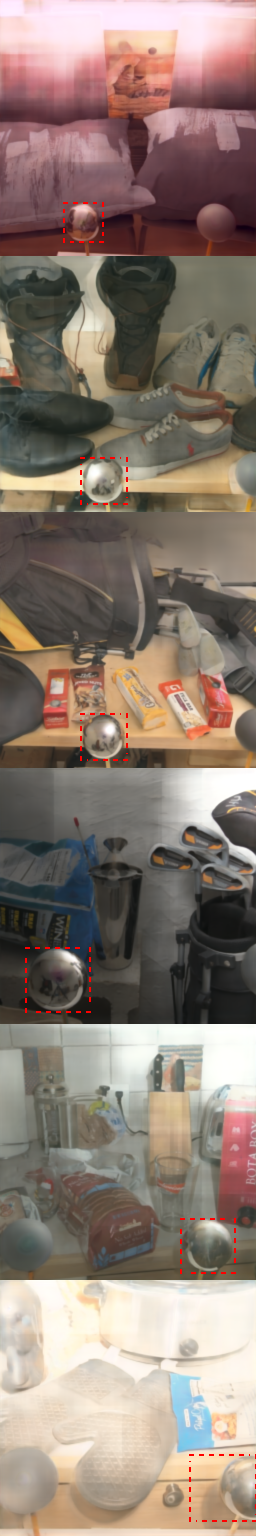}
    \end{minipage}
    \begin{minipage}{0.07\linewidth}
        \centering
        \vspace{-10pt}
        \textbf{\small S3Net Depth}
        \textcolor{black}{\rule[1ex]{\linewidth}{2pt}}
        \includegraphics[width=1.0\textwidth]{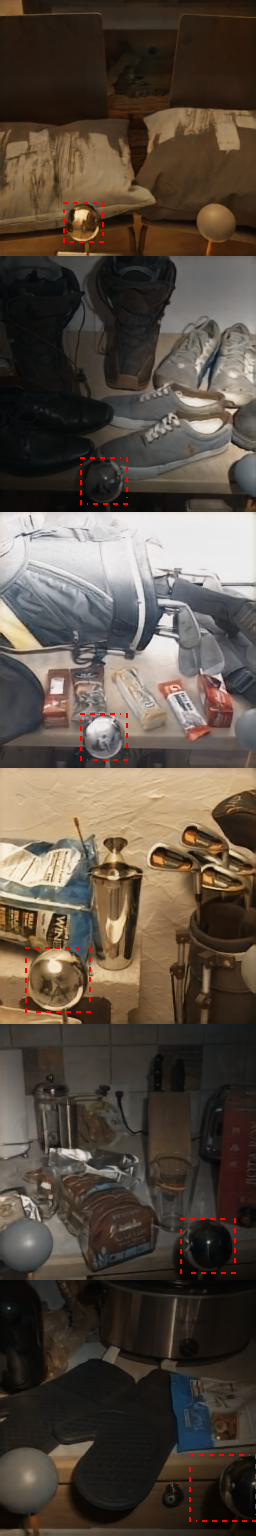}
    \end{minipage}
    \begin{minipage}{0.07\linewidth}
        \vspace{-10pt}
        \centering
        {\small \textbf{Ours unsup}}
        \textcolor{black}{\rule[1ex]{\linewidth}{2pt}}
        \includegraphics[width=1.0\textwidth]{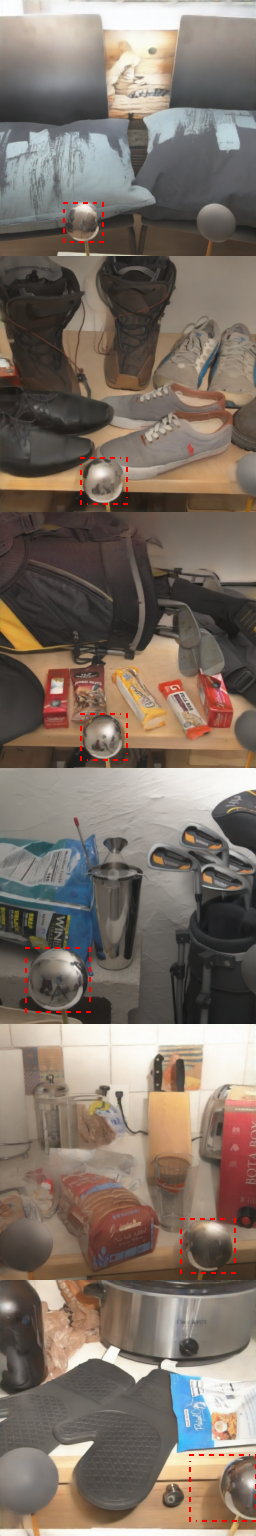}
    \end{minipage}
    \begin{minipage}{0.07\linewidth}
        \centering
        \textbf{Target}
        \textcolor{black}{\rule[1ex]{\linewidth}{2pt}}
        \includegraphics[width=1.0\textwidth]{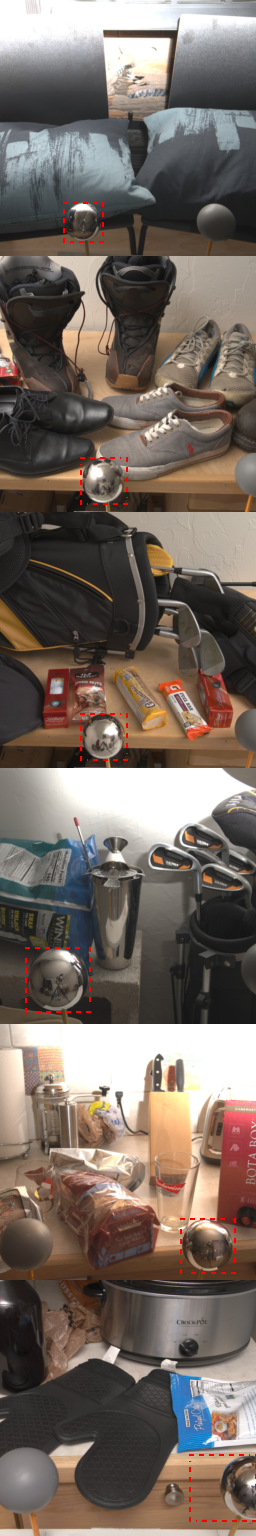}
    \end{minipage}
    \hfill
    \hspace{-2pt}\vline
    \hfill
    \begin{minipage}{0.07\textwidth}
        \centering
        \textbf{Input}
        \textcolor{black}{\rule[1ex]{\linewidth}{2pt}}
        \includegraphics[width=1.0\textwidth]{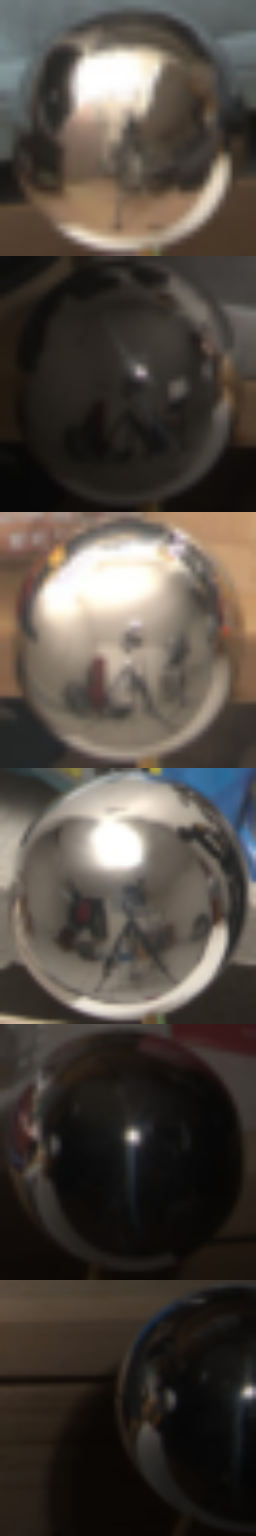}
    \end{minipage}
    \begin{minipage}{0.07\textwidth}
    \vspace{-10pt}
        \centering
        {\small \textbf{SAAE sup}}
        \textcolor{black}{\rule[1ex]{\linewidth}{2pt}}
        \includegraphics[width=1.0\textwidth]{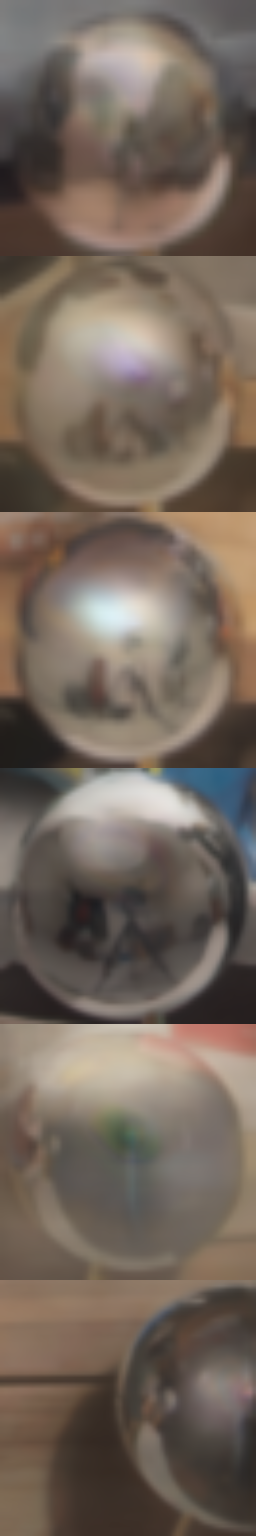}
    \end{minipage}
        \begin{minipage}{0.07\textwidth}
        \vspace{-10pt}
        \centering
        {\small \textbf{SAAE unsup}}
        \textcolor{black}{\rule[1ex]{\linewidth}{2pt}}
        \includegraphics[width=1.0\textwidth]{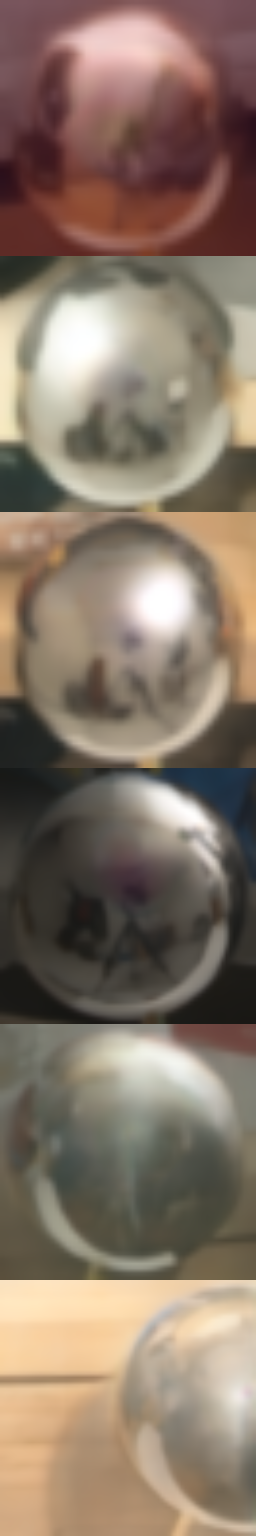}
    \end{minipage}
    \begin{minipage}{0.07\textwidth}
        \centering
        \vspace{-10pt}
        \textbf{\small S3Net Depth}
        \textcolor{black}{\rule[1ex]{\linewidth}{2pt}}
        \includegraphics[width=1.0\textwidth]{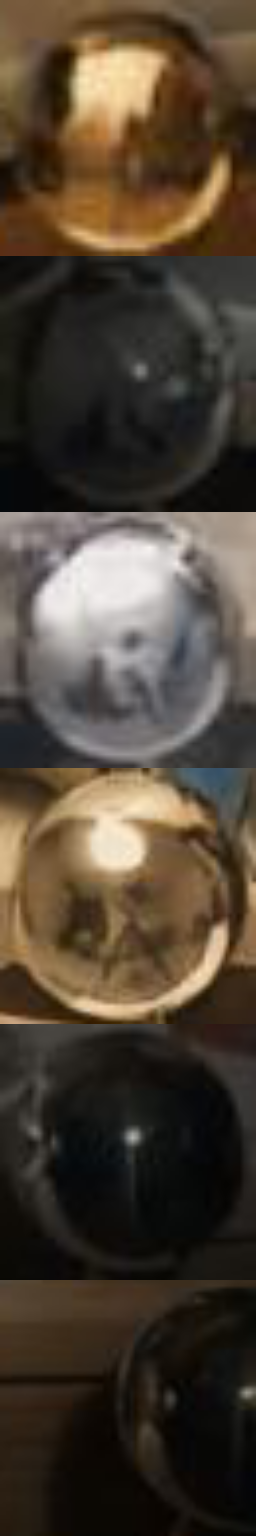}
    \end{minipage}
    \begin{minipage}{0.07\textwidth}
        \vspace{-10pt}
        \centering
       {\small \textbf{Ours unsup}}
        \textcolor{black}{\rule[1ex]{\linewidth}{2pt}}
        \includegraphics[width=1.0\textwidth]{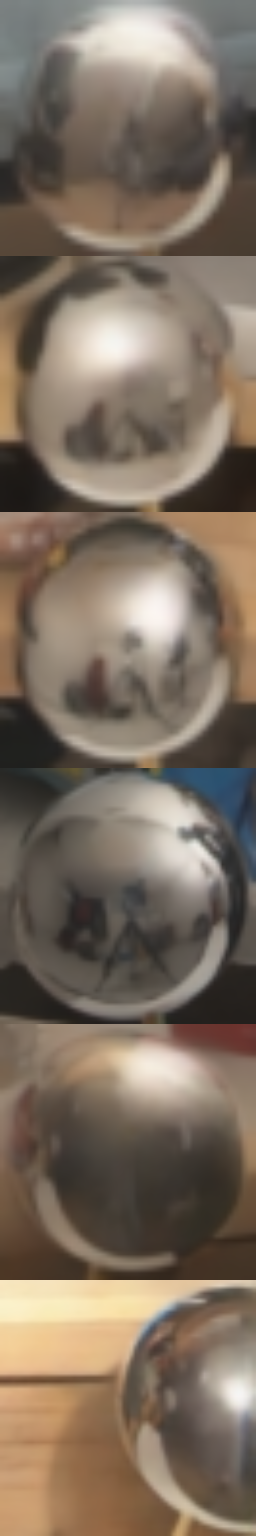}
    \end{minipage}
    \begin{minipage}{0.07\textwidth}
        \centering
        \textbf{Target}
        \textcolor{black}{\rule[1ex]{\linewidth}{2pt}}
        \includegraphics[width=1.0\textwidth]{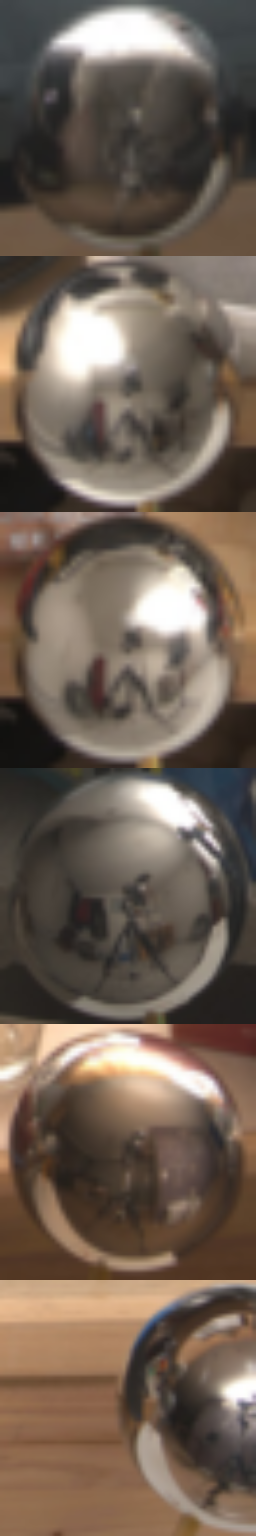}
    \end{minipage}
    \caption{Our method outperforms all other approaches in estimating light and rendering the scene. The Unsupervised SA-AE \cite{hu2020sa} method fails by incorporating intrinsic elements from reference images. The S3Net \cite{yang2021s3net} approach struggles with rendering when using unpaired reference images. \textbf{\textit{Right}}:
    A zoomed-in view of the chrome ball was used as a probe to evaluate detail preservation in the environment map. Our method effectively retains the intricate room layout and accurately renders the appropriate lighting effects.}
    \label{fig:mit relight}
\end{figure}
\begin{table*}[t]
\small
\begin{minipage}[t]{0.58\linewidth}{
\centering
\setlength{\tabcolsep}{5pt}
\scalebox{0.85}{
\begin{tabular}[t]{lc|cc|cc}
\textbf{Methods} & \textbf{Labels} & \multicolumn{2}{c|}{\textbf{Raw Output}} & \multicolumn{2}{c}{\textbf{Color Correction}} \\
\hline
& &  RMSE$\downarrow$ & SSIM$\uparrow$ & RMSE$\downarrow$ & SSIM$\uparrow$ \\
\hline
Input Img & - & 0.384 & 0.438 & 0.312 & 0.492\\
SA-AE~\cite{hu2020sa} & Light & \textbf{0.288} & \textbf{0.484} & 0.232 & 0.559\\
SA-AE~\cite{hu2020sa} & - & 0.443 & 0.300 & 0.317 & 0.431\\
S3Net~\cite{yang2021s3net} & Depth & 0.512 & 0.331 & 0.418 & 0.374 \\
S3Net~\cite{yang2021s3net} & - & 0.499 & 0.336 & 0.414 & 0.377 \\
Ours($\sigma = 0$) & - & 0.326 & 0.232 & 0.242& 0.541\\
Ours(w/o $\mathcal{L}_{reg}$) & - & 0.315 & 0.462	& 0.232 & 0.550\\
Ours & - & 0.297 & 0.473& \textbf{0.222} & \textbf{0.571}\\
\end{tabular}
}}
\end{minipage}
\hfill
\begin{minipage}[t]{0.4\linewidth}{
\setlength{\tabcolsep}{4.0pt}
\scalebox{0.9}{
\begin{tabular}[t]{c|cc|cc}
{$\mathbf{\alpha}$} & \multicolumn{2}{c|}{\textbf{Raw Output}} & \multicolumn{2}{c}{\textbf{Color Correction}} \\
\hline
&  RMSE $\downarrow$ & SSIM $\uparrow$ & RMSE$\downarrow$ & SSIM$\uparrow$ \\
\hline
$\infty$& 0.471 & 0.287 & 0.352 & 0.407 \\
1e-2 & 0.314 & 0.444 & 0.238 & 0.546\\
5e-3 & \textbf{0.297} & \textbf{0.473} & \textbf{0.222} & \textbf{0.571}\\
1e-3 & 0.312 & 0.453 & 0.256 & 0.524\\
5e-4  & 0.309 & 0.460& 0.253& 0.533 \\
\end{tabular}
}}\end{minipage}
\begin{minipage}[t]{0.57\linewidth}
\captionof{table}{We assess the quality of image relighting using the multi-illumination dataset~\citep{murmann2019dataset}. Our method, when evaluated on raw output, significantly outperforms all other unsupervised approaches and achieves competitive results compared to the supervised SA-SA~\citep{hu2020sa}, which requires ground truth light supervision. When we correct the colors by eliminating global color drift caused by light ambiguity, our method surpasses all other approaches. Additionally, warming up the model as a denoising autoencoder proves beneficial compared to when it is not warmed up ($\sigma = 0$).
\label{tab:mit_relight_result}
}
\end{minipage}
\hfill
\hspace{2pt}
\begin{minipage}[t]{0.4\linewidth}
\label{tab:mit_relight_ablation}
\captionof{table}{We analyze the impact of $\alpha$ on relighting quality using the multi-illumination dataset~\citep{murmann2019dataset}. Setting $\alpha$ to $\infty$, which removes the scaling constraints, results in poor relighting quality, indicating that restricting information from extrinsic sources significantly improves generation quality. Within a limited parameter search, 5e-3 yields the best results.} 
\label{tab:alpha_ablation}
\end{minipage}
\vspace{-10pt}
\end{table*}
\textbf{Relighting on the Multi-illumination dataset:}  We relight
images of scenes in the test set using reference images from the test
set, then compare to the correct known relighting from the test set using
various metrics. For each input image, we randomly sample reference images from different scenes and
lighting conditions. To reduce the effect of randomness in comparing
different relighting strategies, we select 12 random
reference images for each input image, and maintain the same
image-reference pairs when evaluating different models. 
We report the results, measured in RMSE and SSIM, in
Table~\ref{tab:mit_relight_result}. 
\begin{figure}
    \centering
    \includegraphics[width = 1.0\textwidth]{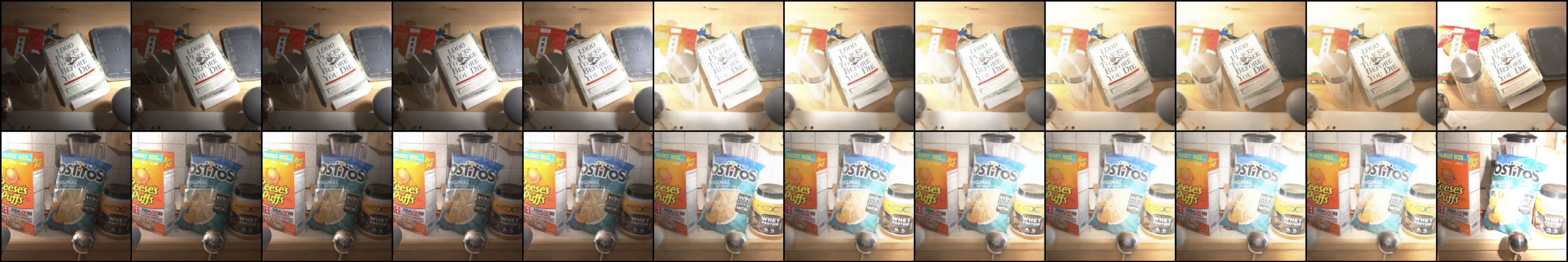}
    \caption{
Latent extrinsics can be interpolated successfully; {\bf leftmost} and {\bf rightmost} columns are images
from the multi-illumination dataset, and intermediate images are obtained by linear interpolation on the
latent extrinsics (light-dependent representations), then decoding.  Note how the light seems to "move" across space.
    }
    \label{fig:interpolation}
    \vspace{-30pt}
\end{figure}
We report these metrics both for absolute predictions and for
predictions where any global color shift is corrected by a single,
least-squares scale of each predicted color layer (i.e. one scale for
R; one for G; one for B). This color correction allows us to
distinguish between spatial errors and global color shifts; these
appear to have a significant effect, possibly because there are
visible color shifts present in some of the dataset images.

In Table~\ref{tab:mit_relight_result}, we compare to SA-AE~\citep{hu2020sa}, a model that requires a ground truth light index for supervision, and S3-Net, which needs a ground truth depth
map as a conditional input. For S3-Net, we use a state-of-the-art
depth estimator to provide pseudo-GT on the relighting dataset as input. For a fair comparison to our
model, which does not require any supervision outside of the ground truth relighting, we also report results for modified versions of the baselines trained without additional supervision. For SA-AE,  we train their light estimation model and relighting model end-to-end by removing the loss from light supervision. For S3-Net~\citep{yang2021s3net}, we simply remove the depth from the model's input. 

Without color correction, only light-supervised SA-AE slightly outperforms our model, while all other baselines are significantly worse. The
unsupervised version of SA-AE performs much worse
because their light estimator struggles to distinguish the extrinsic
from the intrinsic components.  Specifically, SA-AE also parameterizes the extrinsic as a lower-dimensional representation but without the constrained scaling
that our model uses. As a result, the estimated extrinsic from their
unsupervised model also carries intrinsic information, and one can see
``leaks''. S3-Net performs
worse in both versions since they concatenate input and reference
images before feeding them into the models, which significantly
affects the model's generalization ability, especially during test
time when we use images from different scenes as references.  

On color-corrected images, our approach outperforms all methods,
including the light-supervised version of SA-AE, indicating
that, up to the constant color drift, our extrinsic estimation network
is at least as good as, or even better than, a light estimation
network trained with supervision. Removing the denoising setup from
our model ($\sigma=0$) results in worse performance in both cases due
to inferior semantic scene understanding. We additionally provide
ablation studies on the choices of $\alpha$ in Table~\ref{tab:alpha_ablation} and find $\alpha=5e-3$
produces the best results.

Each image in the multi-illumination dataset shows a chrome ball,
which gives a good estimate of an environment map for that image.
Correctly rendering the effects of lighting changes on these chrome
balls appears to be extremely difficult; the changes are substantial,
and concentrated in a small region of the image (so correct
representation of these changes has little effect on typical image losses).
Figure~\ref{fig:mit relight} shows a crop of our results around this
chrome ball. Our method represents these changes well; we are aware of
no other results reported for this effect. Compared to other approaches, our model
accurately preserves the room layout, even in cases of extreme light
changes. 

Unlike classical rendering models that use a specific parameterized form to represent extrinsics, our framework learns an implicit extrinsic representation. However, we can still parameterize the learned extrinsic representation to create new light sources. In Figure~\ref{fig:interpolation}, we demonstrate this capability by rendering images using interpolated extrinsic representations.

\begin{figure}[t]
    \centering
    \begin{minipage}[t]{0.46\linewidth}
    \begin{minipage}{0.24\textwidth}
        \vspace{-10pt}
        \centering
        {\small \textbf{StyleGAN Generation}}
        \textcolor{black}{\rule[1ex]{\linewidth}{2pt}}
        \includegraphics[width=1.0\textwidth]{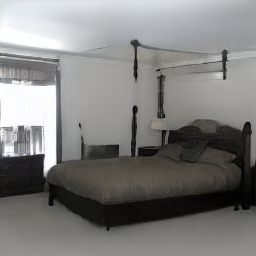}
    \end{minipage}
    \begin{minipage}{0.24\textwidth}
        \centering
        {\small \textbf{Ref}}
        \textcolor{black}{\rule[1ex]{\linewidth}{2pt}}
        \includegraphics[width=1.0\textwidth]{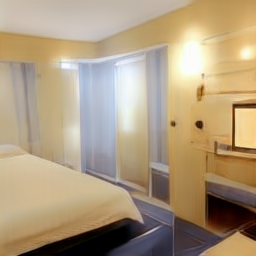}
    \end{minipage}
    \begin{minipage}{0.24\textwidth}
        \centering
        {\small \textbf{Ours}}
        \textcolor{black}{\rule[1ex]{\linewidth}{2pt}}
        \includegraphics[width=1.0\textwidth]{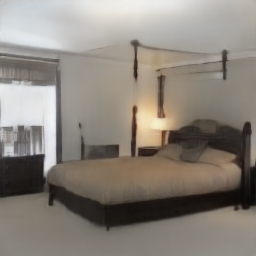}
    \end{minipage}
    \begin{minipage}{0.24\textwidth}
        \vspace{-10pt}
        \centering
        {\small \textbf{StyLitGAN Relight}}
        \textcolor{black}{\rule[1ex]{\linewidth}{2pt}}
        \includegraphics[width=1.0\textwidth]{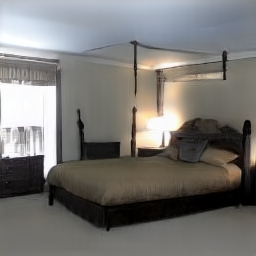}
    \end{minipage}
    \begin{minipage}{0.24\textwidth}
        \centering
        \includegraphics[width=1.0\textwidth]{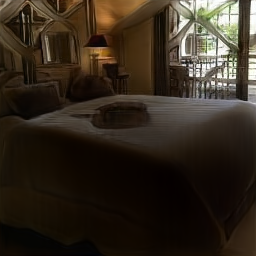}
    \end{minipage}
    \begin{minipage}{0.24\textwidth}
        \centering
        \includegraphics[width=1.0\textwidth]{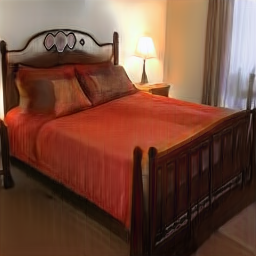}
    \end{minipage}
    \begin{minipage}{0.24\textwidth}
        \centering
        \includegraphics[width=1.0\textwidth]{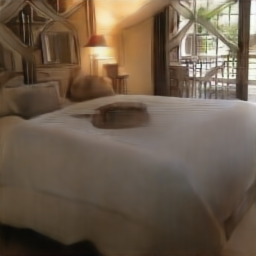}
    \end{minipage}
    \begin{minipage}{0.24\textwidth}
        \centering
        \includegraphics[width=1.0\textwidth]{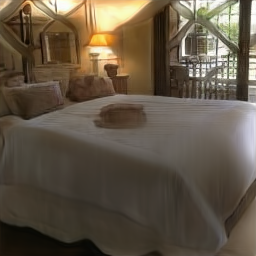}
    \end{minipage}
    
    \vspace{5pt}
      \begin{minipage}{0.24\textwidth}
        \centering
        \includegraphics[width=1.0\textwidth]{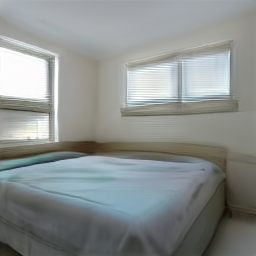}
    \end{minipage}
      \begin{minipage}{0.24\textwidth}
        \centering
        \includegraphics[width=1.0\textwidth]{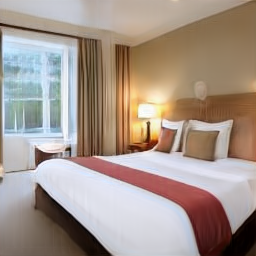}
    \end{minipage}
      \begin{minipage}{0.24\textwidth}
        \centering
        \includegraphics[width=1.0\textwidth]{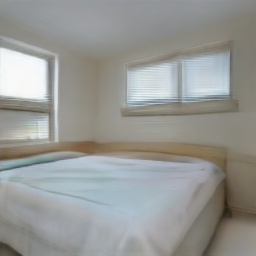}
    \end{minipage}
      \begin{minipage}{0.24\textwidth}
        \centering
        \includegraphics[width=1.0\textwidth]{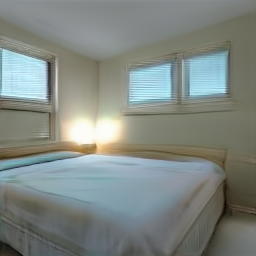}
    \end{minipage}
    \end{minipage}
    \hfill
    \begin{minipage}[t]{0.46\linewidth}
    \begin{minipage}{0.24\textwidth}
        \vspace{-10pt}
        \centering
        {\small \textbf{StyleGAN Generation}}
        \textcolor{black}{\rule[1ex]{\linewidth}{2pt}}
        \includegraphics[width=1.0\textwidth]{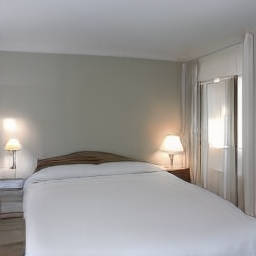}
    \end{minipage}
    \begin{minipage}{0.24\textwidth}
        \centering
        {\small \textbf{Ref}}
        \textcolor{black}{\rule[1ex]{\linewidth}{2pt}}
        \includegraphics[width=1.0\textwidth]{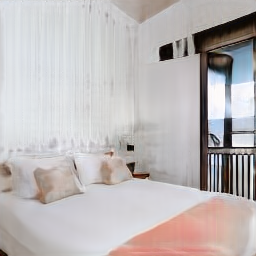}
    \end{minipage}
    \begin{minipage}{0.24\textwidth}
        \centering
        {\small \textbf{Ours}}
        \textcolor{black}{\rule[1ex]{\linewidth}{2pt}}
        \includegraphics[width=1.0\textwidth]{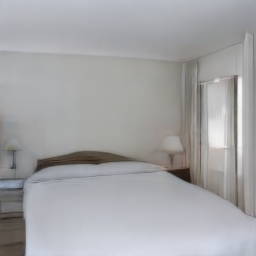}
    \end{minipage}
    \begin{minipage}{0.24\textwidth}
        \vspace{-10pt}
        \centering
        {\small \textbf{StyLitGAN Relight}}
        \textcolor{black}{\rule[1ex]{\linewidth}{2pt}}
        \includegraphics[width=1.0\textwidth]{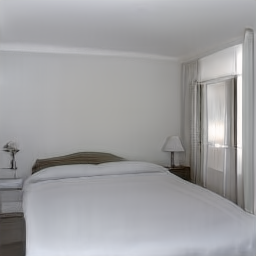}
    \end{minipage}
      \begin{minipage}{0.24\textwidth}
        \centering
        \includegraphics[width=1.0\textwidth]{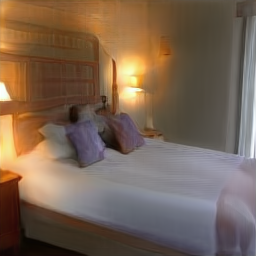}
    \end{minipage}
      \begin{minipage}{0.24\textwidth}
        \centering
        \includegraphics[width=1.0\textwidth]{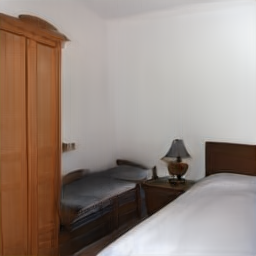}
    \end{minipage}
      \begin{minipage}{0.24\textwidth}
        \centering
        \includegraphics[width=1.0\textwidth]{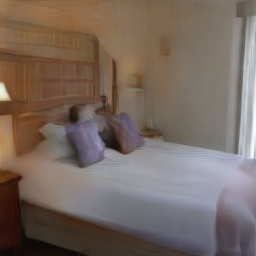}
    \end{minipage}
      \begin{minipage}{0.24\textwidth}
        \centering
        \includegraphics[width=1.0\textwidth]{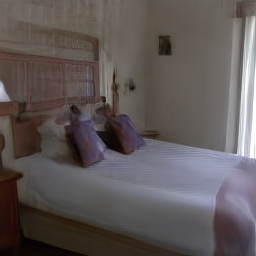}
    \end{minipage}
      \begin{minipage}{0.24\textwidth}
        \centering
        \includegraphics[width=1.0\textwidth]{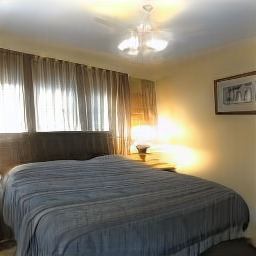}
    \end{minipage}
      \begin{minipage}{0.24\textwidth}
        \centering
        \includegraphics[width=1.0\textwidth]{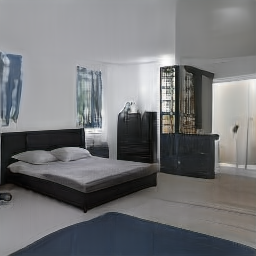}
    \end{minipage}
      \begin{minipage}{0.24\textwidth}
        \centering
        \includegraphics[width=1.0\textwidth]{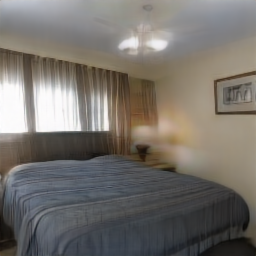}
    \end{minipage}
      \begin{minipage}{0.24\textwidth}
        \centering
        \includegraphics[width=1.0\textwidth]{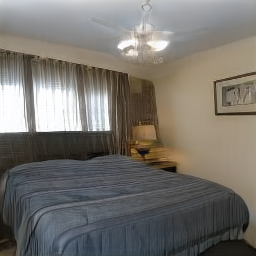}
    \end{minipage}
    \end{minipage}
    \caption{Qualitative results for relighting interior scenes using our relighter
trained on images obtained from StyLitGAN (which produces multiple
illuminations of a generated scene).  StyLitGAN has a strong tendency
to increase or decrease illumination by adjusting luminaires,
typically bedside lights but also light coming through French windows,
etc.  On the {\bf left}, where the reference lighting tends to be
brighter and more concentrated, notice how for the two top images, our
relighter has identified and "turned up" the bedside lights; for the
third, it has resisted StyLitGAN's tendency to invent helpful
luminaires (there isn't a bedside light where StyLitGAN imputed one,
as close inspection shows).  On the {\bf right}, where the
reference lighting is much more uniform, our relighter has achieved
this by "turning down" bedside lights.   This is an emergent
phenomenon; the method is not supplied with any explicit luminaire
model or labeled data. }
    \label{fig:stylegan}
\end{figure}

\textbf{Relighting synthetically relighted images from StyLitGAN:}
StyLitGAN~\citep{StyLitGAN} is a recent method that can produce
multiple illuminations of a single generated room scene by
manipulating StyleGAN latents appropriately.  In the
multi-illumination dataset, reference light and target images tend to share a strong spatial correlation
in light patterns.  In contrast, StyLitGAN generates extremely
challenging images where very significant changes in lighting occur.
Furthermore, StyLitGAN images have visible luminaires.
To relight the input, the model must infer
high-level concepts rather than simply copying the spatially
corresponding light patterns from the reference.  
We train our model using StyLitGAN images to evaluate generalization
qualitatively (quantitative evaluation would be of dubious value, because StyLitGAN
images are generated rather than real).
Figure \ref{fig:stylegan} shows results.  Notice how our method
successfully relights from references, achieves brighter illuminations
by turning on luminaires (here bedside lights), achieves darker scenes
by turning off luminaires, and is somewhat less inclined to invent
luminaires than StyLitGAN is.  The model knows that
light must come from somewhere, and how the effects of light are distributed.
\begin{figure}[t!]
    \begin{minipage}[t]{0.325\linewidth}
    \begin{minipage}{0.31\textwidth}
        \centering
        {\small \textbf{Input}}
        \textcolor{black}{\rule[1ex]{\linewidth}{2pt}}
        \includegraphics[width=1.0\textwidth]{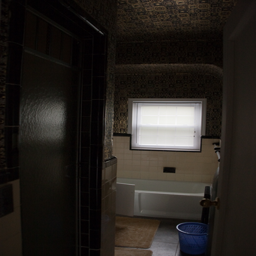}
    \end{minipage}
    \begin{minipage}{0.31\textwidth}
        \centering
        {\small \textbf{Ref}}
        \textcolor{black}{\rule[1ex]{\linewidth}{2pt}}
        \includegraphics[width=1.0\textwidth]{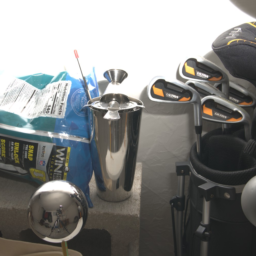}
    \end{minipage}
    \begin{minipage}{0.31\textwidth}
        \centering
         {\small \textbf{Relight}}
        \textcolor{black}{\rule[1ex]{\linewidth}{2pt}}
        \includegraphics[width=1.0\textwidth]{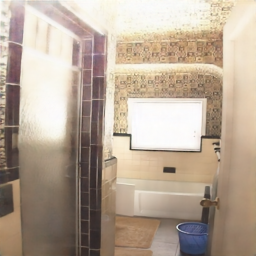}
    \end{minipage}
    \end{minipage}
    \hfill
     \begin{minipage}[t]{0.325\linewidth}
    \begin{minipage}{0.31\textwidth}
        \centering
         {\small \textbf{Input}}
        \textcolor{black}{\rule[1ex]{\linewidth}{2pt}}
        \includegraphics[width=1.0\textwidth]{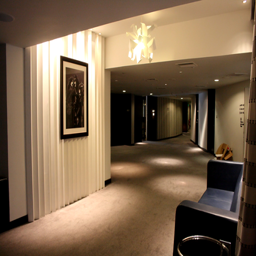}
    \end{minipage}
    \begin{minipage}{0.31\textwidth}
        \centering
        {\small \textbf{Ref}}
        \textcolor{black}{\rule[1ex]{\linewidth}{2pt}}
        \includegraphics[width=1.0\textwidth]{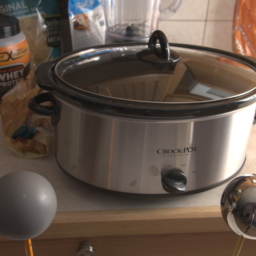}
    \end{minipage}
    \begin{minipage}{0.31\textwidth}
        \centering
        {\small \textbf{Relight}}
        \textcolor{black}{\rule[1ex]{\linewidth}{2pt}}
        \includegraphics[width=1.0\textwidth]{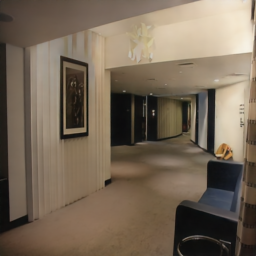}
    \end{minipage}
    \end{minipage}
   \hfill
    \begin{minipage}[t]{0.325\linewidth}
    \begin{minipage}{0.31\textwidth}
        \centering
       {\small \textbf{Input}}
        \textcolor{black}{\rule[1ex]{\linewidth}{2pt}}
        \includegraphics[width=1.0\textwidth]{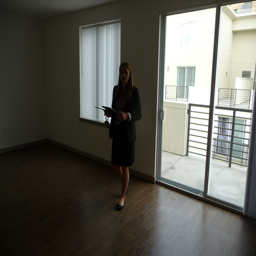}
    \end{minipage}
    \begin{minipage}{0.31\textwidth}
        \centering
       {\small \textbf{Ref}}
        \textcolor{black}{\rule[1ex]{\linewidth}{2pt}}
        \includegraphics[width=1.0\textwidth]{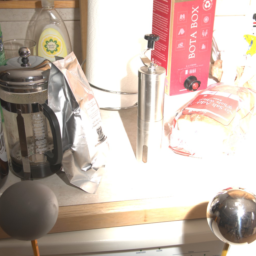}
    \end{minipage}
    \begin{minipage}{0.31\textwidth}
        \centering
       {\small \textbf{Relight}}
        \textcolor{black}{\rule[1ex]{\linewidth}{2pt}}
        \includegraphics[width=1.0\textwidth]{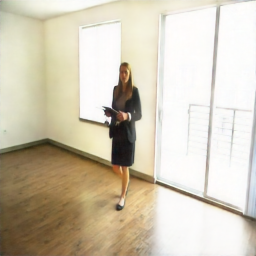}
    \end{minipage}
    \end{minipage}
    \hfill
     \begin{minipage}[t]{0.325\linewidth}
    \begin{minipage}{0.31\textwidth}
        \centering
        \includegraphics[width=1.0\textwidth]{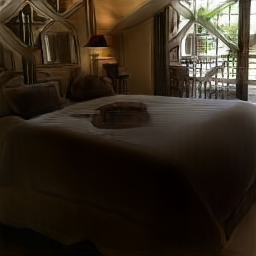}
    \end{minipage}
    \begin{minipage}{0.31\textwidth}
        \centering
        \includegraphics[width=1.0\textwidth]{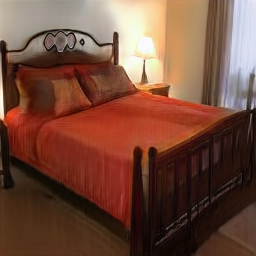}
    \end{minipage}
    \begin{minipage}{0.31\textwidth}
        \centering
        \includegraphics[width=1.0\textwidth]{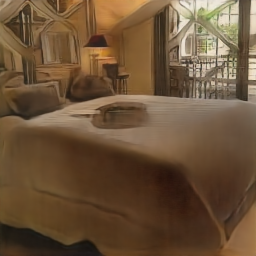}
    \end{minipage}
    \end{minipage}
   \hfill
     \begin{minipage}[t]{0.325\linewidth}
    \begin{minipage}{0.31\textwidth}
        \centering
        \includegraphics[width=1.0\textwidth]{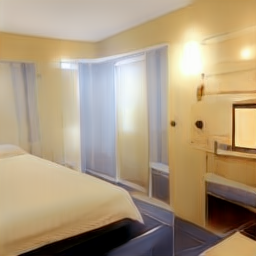}
    \end{minipage}
    \begin{minipage}{0.31\textwidth}
        \centering
        \includegraphics[width=1.0\textwidth]{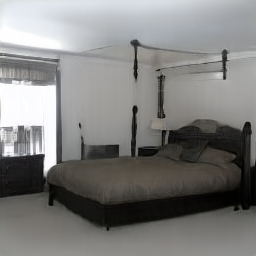}
    \end{minipage}
    \begin{minipage}{0.31\textwidth}
        \centering
        \includegraphics[width=1.0\textwidth]{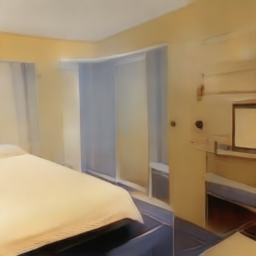}
    \end{minipage}
    \end{minipage}
    \hfill
     \begin{minipage}[t]{0.325\linewidth}
    \begin{minipage}{0.31\textwidth}
        \centering
        \includegraphics[width=1.0\textwidth]{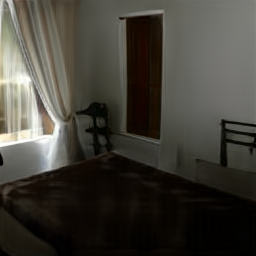}
    \end{minipage}
    \begin{minipage}{0.31\textwidth}
        \centering
        \includegraphics[width=1.0\textwidth]{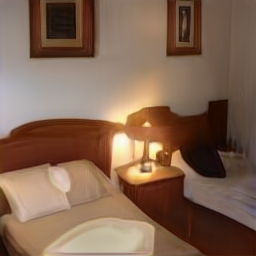}
    \end{minipage}
    \begin{minipage}{0.31\textwidth}
        \centering
        \includegraphics[width=1.0\textwidth]{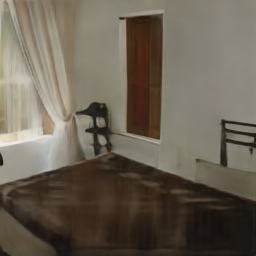}
    \end{minipage}
    \end{minipage}
    \caption{\textbf{Zero-Shot Relighting.} Our relighting model, trained only on the multi-illumination dataset, generalizes well to out-of-distribution images, as shown on the IIW dataset (first row) and StyleGAN images (second row). It accurately infers scene geometry and lighting. Note that it identifies and turns on the bedside lamps in StyleGAN images despite having no training in bedroom images. This demonstrates the model's strong generalization ability and the model clearly ``knows" something about light sources.
    }
    \label{fig:relight_out}
\end{figure}

\textbf{Zero-Shot Relighting:} 
In Figure.\ref{fig:relight_out}, we show our model's strong generalization by applying the model solely trained on multi-illumination dataset—without additional training or fine-tuning—to relight IIW and StyleGAN-generated images. Despite the significant distribution shift in lighting patterns and room setup, our model accurately identifies luminaires and relights images.

\begin{table*}[t]
\small
\begin{minipage}[t]{0.5\linewidth}
\centering
\setlength{\tabcolsep}{4pt}
\begin{tabular}[t]{l|ccc|c}
Methods &    labels  &  Flat & Tune $\delta$ & WHDR \\
\hline
Intrinsic Diffusion \cite{kocsis2023intrinsic} & CG & No & No  & 22.61 \\
Intrinsic Diffusion\cite{kocsis2023intrinsic} & CG & Yes & Yes & 17.10 \\
Inverser Render\cite{yu2019inverserendernet} & No & No & No & 21.40\\
BBA\cite{forsyth2021intrinsic} & No & No & Yes & 17.04 \\
Ours & No &  No & No & 28.97 \\
Ours & No &  No & Yes & 19.09 \\
Ours & No &  Yes & Yes & \textbf{15.81}\\
\end{tabular}
\end{minipage}
\hfill
\begin{minipage}[t]{0.4\linewidth}
\begin{tabular}[t]{c|cc|cc}
$\alpha$  & \multicolumn{3}{c}{WHDR} \\
 \hline
 & \multicolumn{2}{c}{$\delta$ = 0.1}&\multicolumn{2}{c}{optimal $\delta$} \\
 & w/ F & w/o F & w/ F & w/o F \\
 \hline
 1e-2  & \textbf{17.64}  & \textbf{28.97} & \textbf{15.81} & \textbf{19.09} \\
 5e-3 &  18.93 & 31.81 & 16.02 & 19.53 \\
 1e-3 & 18.00 & 29.77 &  15.84
 & 19.13 \\
 5e-4 & 18.04 & 29.62  & 15.85 & 19.12  
\end{tabular}
\end{minipage}
\begin{minipage}[t]{0.56\linewidth}
\vspace{10pt}
   \captionof{table}{We benchmark our albedo esimation on test set of IIW dataset~\citep{bell14intrinsic_short} and compare with others, though the reliability has been questioned by recent papers~\citep{forsyth2021intrinsic}. Flat denotes postprocessing images with flattening~\citep{bi20151}. Despite our model never being trained on albedo maps or CG data, our best configuration significantly outperforms all other methods suggesting our model learns high-quality intrinsic representations} 
   \label{tab:iiw}
\end{minipage}
\hfill
\begin{minipage}[t]{0.4\linewidth}
   \captionof{table}{We conduct ablation experiments to assess the impact of $\alpha$ on the quality of albedo. "w/F" and "w/o F" denote post-processing images with and without flattening~\citep{bi20151}, respectively. The setting of $\delta=0.1$ and w/o F is the most affected by $\alpha$. Despite this, all values of $\alpha$ achieve high performance in our optimal configurations.} 
\end{minipage}
\label{tab:oracle_decoding}
\end{table*}
\begin{figure}[t]
    \centering
    \begin{minipage}{0.09\linewidth}
        \centering
        {\small \textbf{Input}}
        \textcolor{black}{\rule[1ex]{\linewidth}{2pt}}
        \includegraphics[width=1.0\textwidth]{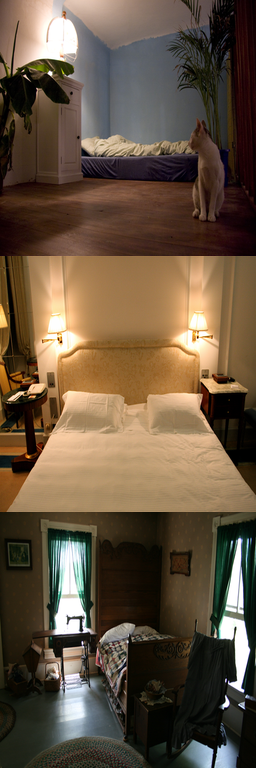}
    \end{minipage}
    \begin{minipage}{0.09\linewidth}
        \vspace{-10pt}
        \centering
        {\small \textbf{Input + Flatten}}
        \textcolor{black}{\rule[1ex]{\linewidth}{2pt}}
        \includegraphics[width=1.0\textwidth]{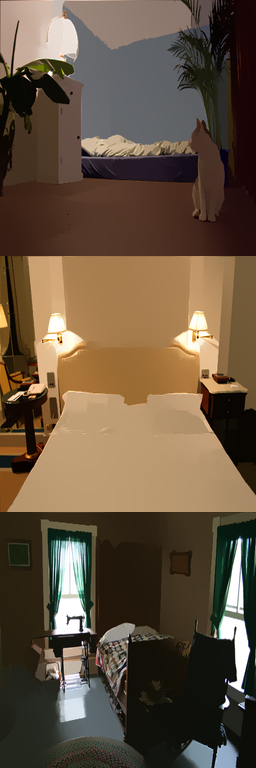}
    \end{minipage}
    \begin{minipage}{0.09\linewidth}
        \centering
        {\small \textbf{Ours}}
        \textcolor{black}{\rule[1ex]{\linewidth}{2pt}}
        \includegraphics[width=1.0\textwidth]{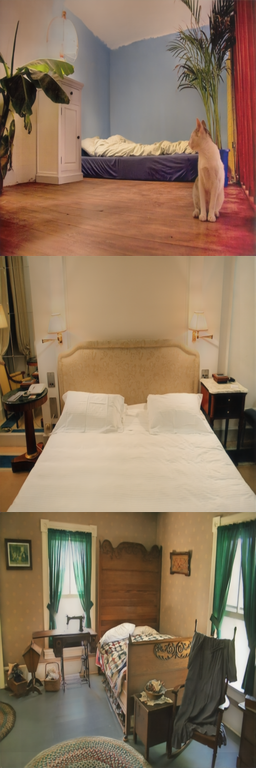}
    \end{minipage}
    \begin{minipage}{0.09\linewidth}
        \vspace{-10pt}
        \centering
        {\small \textbf{Ours + Flatten}}
        \textcolor{black}{\rule[1ex]{\linewidth}{2pt}}
        \includegraphics[width=1.0\textwidth]{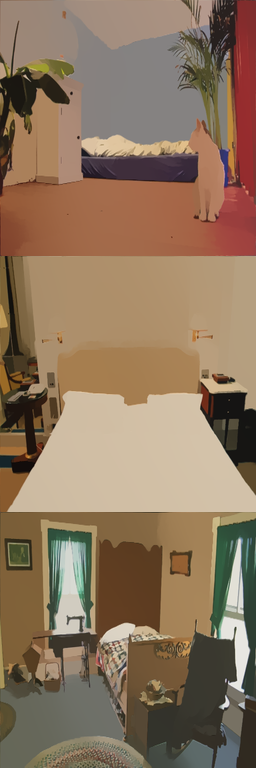}
    \end{minipage}
    \begin{minipage}{0.09\linewidth}
        \vspace{-10pt}
        \centering
        {\small \textbf{Intrinsic Diffusion}}
        \textcolor{black}{\rule[1ex]{\linewidth}{2pt}}
        \includegraphics[width=1.0\textwidth]{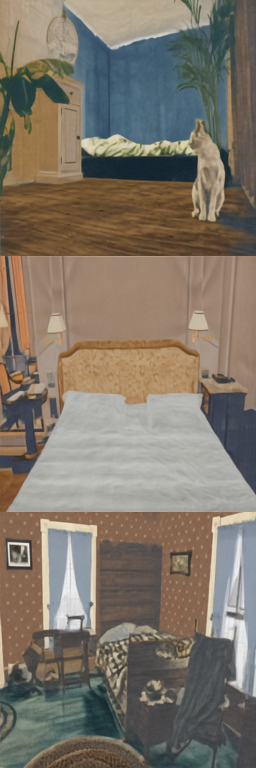}
    \end{minipage}
    \hfill
    \begin{minipage}{0.09\linewidth}
        \centering
        {\small \textbf{Input}}
        \textcolor{black}{\rule[1ex]{\linewidth}{2pt}}
        \includegraphics[width=1.0\textwidth]{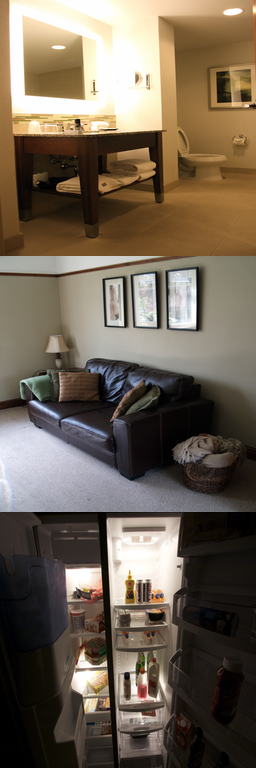}
    \end{minipage}
    \begin{minipage}{0.09\linewidth}
        \vspace{-10pt}
        \centering
        {\small \textbf{Input + Flatten}}
        \textcolor{black}{\rule[1ex]{\linewidth}{2pt}}
        \includegraphics[width=1.0\textwidth]{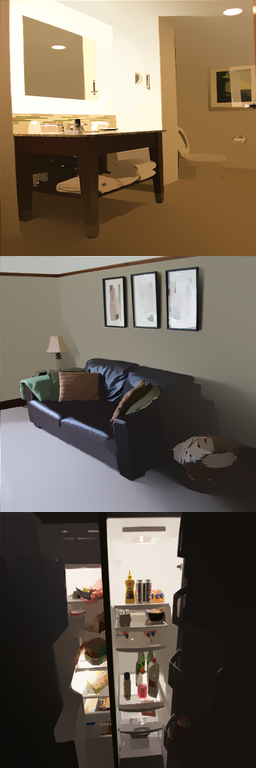}
    \end{minipage}
    \begin{minipage}{0.09\linewidth}
        \centering
        {\small \textbf{Ours}}
        \textcolor{black}{\rule[1ex]{\linewidth}{2pt}}
        \includegraphics[width=1.0\textwidth]{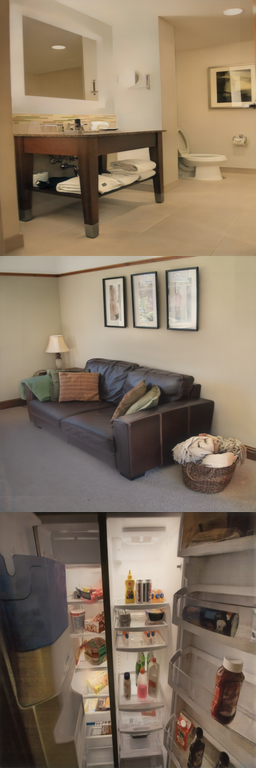}
    \end{minipage}
    \begin{minipage}{0.09\linewidth}
        \vspace{-10pt}
        \centering
        {\small \textbf{Ours + Flatten}}
        \textcolor{black}{\rule[1ex]{\linewidth}{2pt}}
        \includegraphics[width=1.0\textwidth]{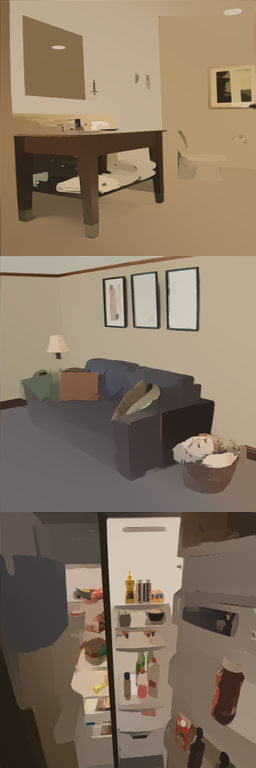}
    \end{minipage}
    \begin{minipage}{0.09\linewidth}
        \vspace{-10pt}
        \centering
        {\small \textbf{Intrinsic Diffusion}}
        \textcolor{black}{\rule[1ex]{\linewidth}{2pt}}
        \includegraphics[width=1.0\textwidth]{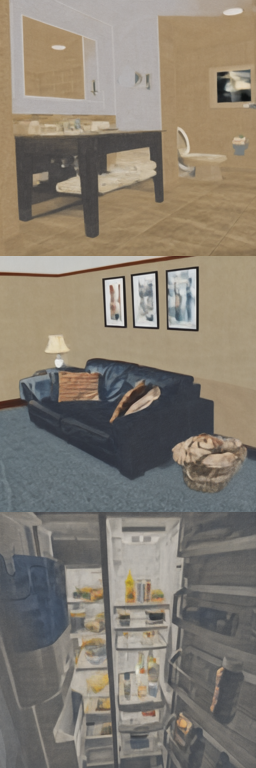}
    \end{minipage} 
    \caption{
Qualitative Comparison of \textbf{Emergent Albedo from Latent Intrinsics} on the IIW Dataset. Although our model has never been trained on any albedo-like maps, it effectively removes the effects of external light and dark shadows from the input. In contrast, Intrinsic Diffusion \citep{kocsis2023intrinsic}, a supervised method trained on large computer graphics data, often produces color-drifted estimations, likely due to the domain shift between CG data and real images. Observe the subdued lighting around the mirrors (top row, right) in our recovered albedo. Also, pay attention to all the details inside the refrigerator, which are visible in our recovered albedos (bottom row; right) compared to intrinsic diffusion. For comparison, we also display naive flattening (in the second column), which by itself cannot effectively reduce the strong lighting effects.}
    \label{fig:albedo_estimation}
    \vspace{-10pt}
\end{figure}
\subsection{Zero-shot albedo evaluation}
\label{sec:albedo_eval}
Constrained scaling allows us to infer albedo without any decoding
(and without any albedo data!) by setting $\alpha=0$ during
inference. We benchmark these albedo estimates using the WHDR metric
on the IIW~\citep{bell14intrinsic_short} dataset
(Section~\ref{sec:related}).
We use WHDR because it is widely used and allows comparisons,
but existing literature records significant problems in interpreting the
measure~\citep{forsyth2021intrinsic, bhattad2022cut,
  kocsis2023intrinsic}. Among other irritating features, the
metric seems to prefer odd colors, and can be hacked by heavily quantized albedo maps.
As is standard, we obtain lightness by averaging R, G, and B albedo and compute relative lightness of two  pixel
locations ${i_1, i_2}$ by comparing to a
confidence threshold $\delta$: 
\begin{eqnarray}
   \widetilde{J}_{i, \delta}(
   \widebar{R}) = \begin{Bmatrix}
   &1& \text{  if  }\widebar{R}_{i_1} / \widebar{R}_{i_2} > 1 + \delta \\
   &2& \text{  if  }\widebar{R}_{i_2} / \widebar{R}_{i_1} > 1 + \delta \\
   &E& \text{  otherwise  } 
   \end{Bmatrix}
\end{eqnarray}
The resulting classification (one lighter than two; two lighter than
one; equivalent) is then compared to  human annotations $J$ using the confidence score $w_i$ for each annotation pair.
We report WHDR on the IIW test split in Table
\ref{tab:iiw} to facilitate comparison with other approaches. Since
our model is not trained with any albedo maps or computer-generated
images, we need to adjust the threshold for the optimal
performance. Following prior work, we optimize $\delta$ on the
training split, which significantly improves our performance from
28.97 to 19.09. Additionally, we enhance our performance by
post-processing our albedo map using flattening~\citep{bi20151}, an
optimization technique to further reduce color variations. With this
improvement, our results reach 15.81, substantially outperforming the
intrinsic diffusion model ~\citep{kocsis2023intrinsic}, a
diffusion-based albedo regression model trained on computer graphics
data. In Figure~\ref{fig:albedo_estimation}, we show some qualitative comparisons to intrinsic diffusion. We observe that our method effectively removes external lighting effects and does not suffer from color drift due to domain gap unlike intrinsic diffusion, which is trained on CG data.
\begin{figure}
    \centering
    \begin{minipage}[t]{0.05\linewidth}%
      \vspace{-10pt}
    \begin{sideways}\scriptsize{\textbf{\textsf{Ours~~~~~~~~Intrinsic~~~~~~~~Image}}}\end{sideways}
    \begin{sideways}\scriptsize{\textbf{\textsf{~~~~~~~~~~~~~~~~Diffusion}}}\end{sideways}
   \end{minipage}%
    \begin{minipage}[t]{0.93\linewidth}%
    \begin{minipage}[t]{0.48\linewidth}
    \begin{minipage}{0.23\textwidth}
        \centering
        \includegraphics[width=1.0\textwidth]{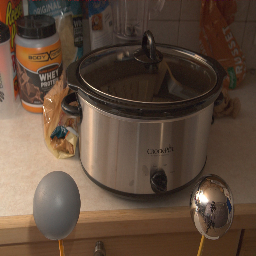}
    \end{minipage}
    \begin{minipage}{0.23\textwidth}
        \centering
        \includegraphics[width=1.0\textwidth]{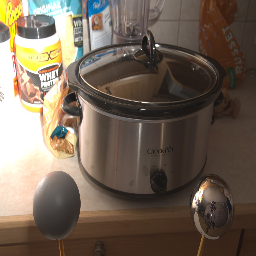}
    \end{minipage}
     \begin{minipage}{0.23\textwidth}
        \centering
        \includegraphics[width=1.0\textwidth]{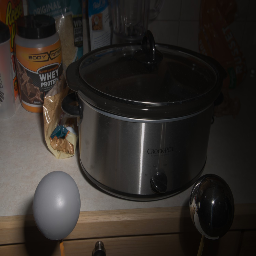}
    \end{minipage} 
     \begin{minipage}{0.23\textwidth}
        \centering
        \includegraphics[width=1.0\textwidth]{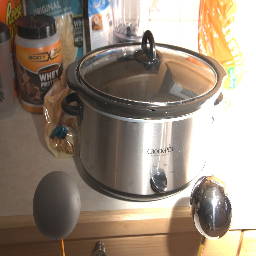}
    \end{minipage} 

    \begin{minipage}{0.23\textwidth}
        \centering
        \includegraphics[width=1.0\textwidth]{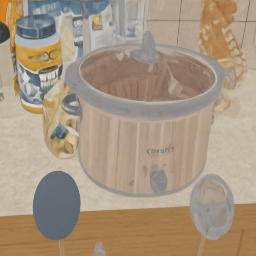}
    \end{minipage}
    \begin{minipage}{0.23\textwidth}
        \centering
        \includegraphics[width=1.0\textwidth]{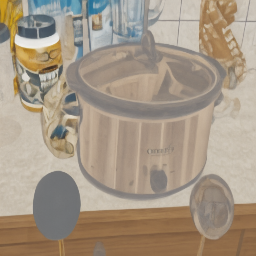}
    \end{minipage}
     \begin{minipage}{0.23\textwidth}
        \centering
        \includegraphics[width=1.0\textwidth]{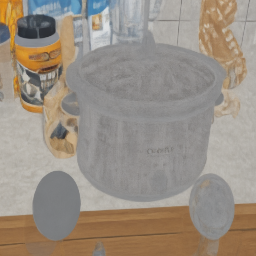}
    \end{minipage} 
     \begin{minipage}{0.23\textwidth}
        \centering
        \includegraphics[width=1.0\textwidth]{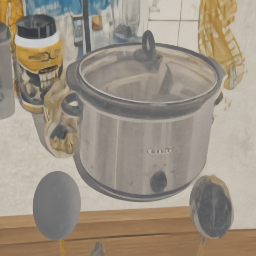}
    \end{minipage} 
    
    \begin{minipage}{0.23\textwidth}
        \centering
        \includegraphics[width=1.0\textwidth]{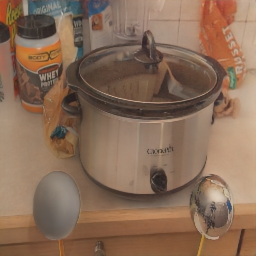}
    \end{minipage}
    \begin{minipage}{0.23\textwidth}
        \centering
        \includegraphics[width=1.0\textwidth]{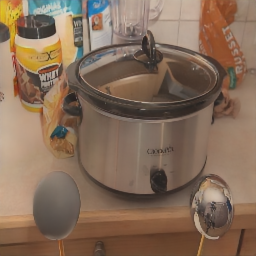}
    \end{minipage}
     \begin{minipage}{0.23\textwidth}
        \centering
        \includegraphics[width=1.0\textwidth]{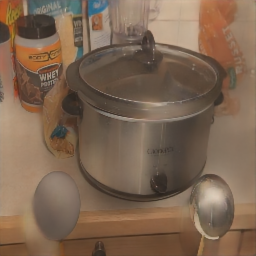}
    \end{minipage} 
     \begin{minipage}{0.23\textwidth}
        \centering
        \includegraphics[width=1.0\textwidth]{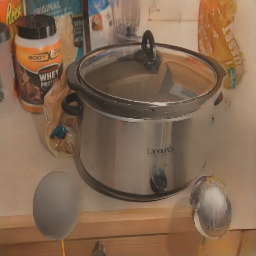}
    \end{minipage} 
    
    \end{minipage}
    \hfill
    \begin{minipage}[t]{0.48\linewidth}
    \begin{minipage}{0.23\textwidth}
        \centering
        \includegraphics[width=1.0\textwidth]{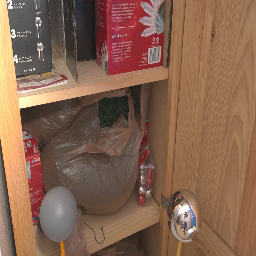}
    \end{minipage}
    \begin{minipage}{0.23\textwidth}
        \centering
        \includegraphics[width=1.0\textwidth]{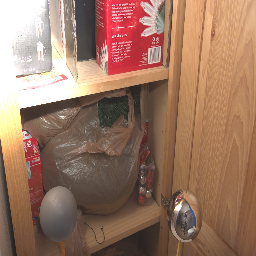}
    \end{minipage}
     \begin{minipage}{0.23\textwidth}
        \centering
        \includegraphics[width=1.0\textwidth]{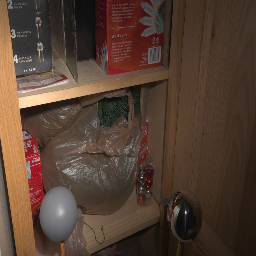}
    \end{minipage} 
     \begin{minipage}{0.23\textwidth}
        \centering
        \includegraphics[width=1.0\textwidth]{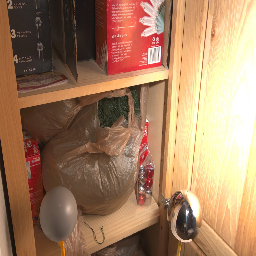}
    \end{minipage} 

    \begin{minipage}{0.23\textwidth}
        \centering
        \includegraphics[width=1.0\textwidth]{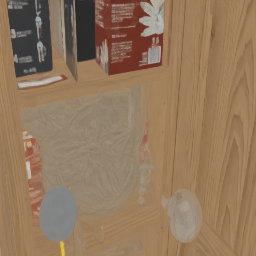}
    \end{minipage}
    \begin{minipage}{0.23\textwidth}
        \centering
        \includegraphics[width=1.0\textwidth]{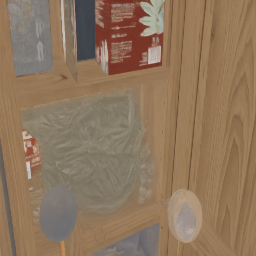}
    \end{minipage}
     \begin{minipage}{0.23\textwidth}
        \centering
        \includegraphics[width=1.0\textwidth]{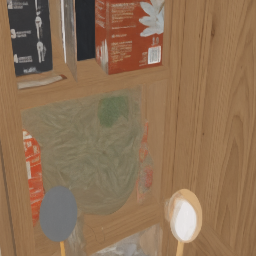}
    \end{minipage} 
     \begin{minipage}{0.23\textwidth}
        \centering
        \includegraphics[width=1.0\textwidth]{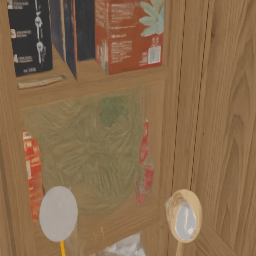}
    \end{minipage} 
    
    \begin{minipage}{0.23\textwidth}
        \centering
        \includegraphics[width=1.0\textwidth]{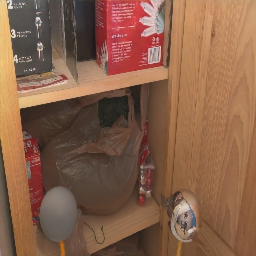}
    \end{minipage}
    \begin{minipage}{0.23\textwidth}
        \centering
        \includegraphics[width=1.0\textwidth]{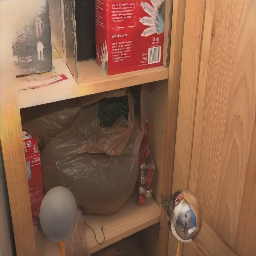}
    \end{minipage}
     \begin{minipage}{0.23\textwidth}
        \centering
        \includegraphics[width=1.0\textwidth]{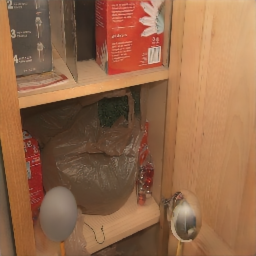}
    \end{minipage} 
     \begin{minipage}{0.23\textwidth}
        \centering
        \includegraphics[width=1.0\textwidth]{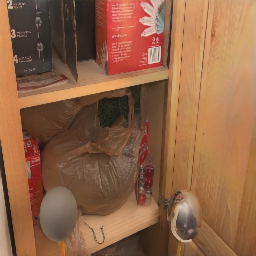}
    \end{minipage} 
    \end{minipage}
   \end{minipage}%
\caption{Qualitative comparison of albedo stability under varying lighting conditions. Images shown are from the multi-illumination dataset test split. The top row features images under different lighting environments. The middle row presents estimated albedos obtained from Intrinsic Diffusion~\cite{kocsis2023intrinsic}, while the bottom row shows the recovered albedos from the latent intrinsic representation. Intrinsic Diffusion has large color drift and is sensitive to changes in lighting. In contrast, the \textbf{albedos recovered from latent intrinsics remain stable under lighting changes, even in extreme conditions.}
    }
    \vspace{-5pt}
    \label{fig:relight_sensitivity}
\end{figure}

\textbf{Sensitivity to light changes:} Albedo are scene properties that are independent of lighting changes. In Figure.~\ref{fig:relight_sensitivity}, we qualitatively assess this characteristic by varying lighting conditions, comparing our approach with the state-of-the-art supervised method, Intrinsic Diffusion \citep{kocsis2023intrinsic}. Our method demonstrates consistent and accurate estimations that remain stable even under extreme lighting variations. In contrast, Intrinsic Diffusion \citep{kocsis2023intrinsic} shows significant deviation from the natural color distribution and are sensitive to lighting changes.
\section{Discussion, Limitations and Future Work}
\label{sec:discuss}
Our method presents an important advancement in image relighting by demonstrating that intrinsic properties such as albedo can emerge naturally from training on relighting tasks without explicit supervision. This finding simplifies the relighting process, eliminating the need for detailed geometric and surface models and enhancing the model’s ability to generalize across diverse and unseen scenes. By encoding scene and illumination properties as latent variables, we achieve accurate and flexible relighting. Our findings will have implications for various fields such as virtual reality and cinematic post-production. This approach reduces the learning process's complexity and offers a new perspective on designing deep learning models to capture and utilize intrinsic scene properties. These findings can guide future research toward a more efficient and scalable relighting approach, encouraging the development of models that can handle various lighting conditions and scene complexities.

The current taxonomy of surface intrinsics—typically, depth, normal, albedo, and perhaps specular albedo and roughness—is quite limiting (compare human language for surface properties~\cite{beckbook}). Our method, which computes latent intrinsic and extrinsic representations from images and combines these to transfer lighting conditions across scenes, captures physical concepts like luminaire and albedo without explicit physical parametrization. This ability to represent significant image effects without choosing a surface model offers substantial flexibility.

However, our method has several limitations. It relies on pairs of relighted data captured in the same scene, which can be resource-intensive to obtain. Additionally, it does not cope well with saturated pixel values common in LDR images. The intrinsic information being latent is another limitation since many applications require explicit intrinsic information like depth and normals. 

Nonetheless, there is good evidence that explicit intrinsic information can be extracted from our latent intrinsics. Our method clearly ``knows'' albedo, and this information can be elicited without examples. Similarly, it ``knows'' something about luminaires, such as their locations and effects. It is intriguing to speculate that it ``knows'' other information relevant to relighting, such as depth or surface microstructure. Future work will pursue this line of inquiry and also focus on developing a purely unsupervised framework to infer intrinsic and extrinsic properties from collections of in-the-wild images. This will include refining probing techniques for better extraction of explicit intrinsics and identifying additional intrinsic properties crucial for relighting that do not align with the current taxonomy. We believe this will improve the applicability and robustness of our approach, making it suitable for a wider range of real-world scenarios.

\section*{Acknowledgment}
AB thanks Stephan R. Richter for the discussions that led to the consideration of intrinsic images as latent variables. This material in part is based upon work supported by the National Science Foundation under Grant No. 2106825, and
by a gift from Boeing.
\bibliography{dafsupp,bigdaf,references,main,refs-daf,more,refs}
\bibliographystyle{abbrvnat}

\appendix
\section{Experiment Details}
\label{sec:exp_details}
\textbf{Training Details} We train our model with a batch size of 256 for 1,000 epochs using the AdamW optimizer, with a constant learning rate of 2e-4 and a weight decay ratio of 1e-2. To improve the semantic representation, we corrupt images with Gaussian noise during the first 400 epochs and follow~\citet{karras2022elucidating} to sample the standard deviation $\sigma$ with $\ln(\sigma)\sim \mathcal{N}(-1.2, 1.2^2)$. In the later 600 epochs, we turn off the Gaussian noise to focus on enhancing the image quality. We train our model with 4A40 and a complete training requires 40 hours.

\textbf{Model Details} Our autoencoder employs a U-Net architecture, incorporating residual convolutional blocks as the fundamental components. Each block is composed of two convolutional layers, group normalization, and a nonlinear activation function. The structure specifies [1, 2, 2, 4, 4, 4] blocks at each resolution level, starting from a resolution of 256, with the resolution halving after each level. The corresponding configurations for latent channels at these levels are [32, 64, 128, 128, 256, 512].

The intrinsic features, denoted as ${\mS_{s,i}^l}$,  are gathered from the output of the final block at each resolution level, starting from a resolution of 128x128 down to the bottleneck. For generating extrinsic features ${\mL_s^l}$, multiple MLP layers are applied to the bottleneck features of the encoder, followed by averaging across all spatial features. We limit the channel number of the extrinsic features to 16 to prevent them from conveying high-frequency components.
\begin{figure}[t]
    \centering
    \begin{minipage}{0.11\linewidth}
        \centering
        {\small \textbf{Input}}
        \textcolor{black}{\rule[1ex]{\linewidth}{2pt}}
        \includegraphics[width=1.0\textwidth]{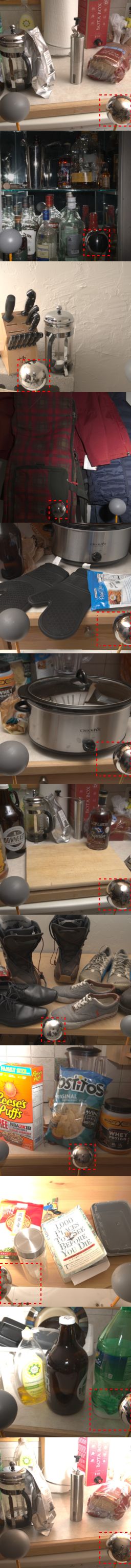}
    \end{minipage}
    \begin{minipage}{0.11\linewidth}
        \centering
        {\small \textbf{Ref}}
        \textcolor{black}{\rule[1ex]{\linewidth}{2pt}}
        \includegraphics[width=1.0\textwidth]{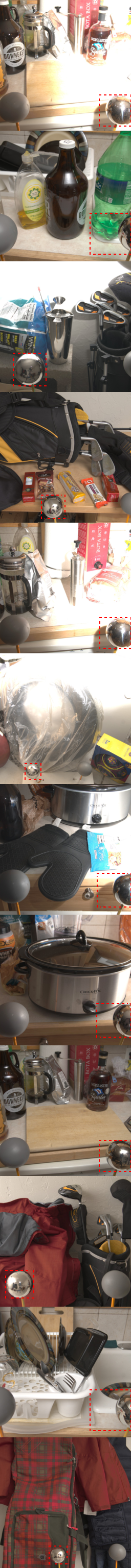}
    \end{minipage}
    \begin{minipage}{0.11\linewidth}
        \centering
        {\small \textbf{Ours}}
        \textcolor{black}{\rule[1ex]{\linewidth}{2pt}}
        \includegraphics[width=1.0\textwidth]{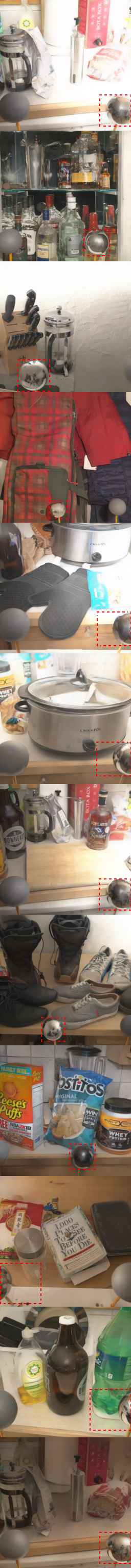}
    \end{minipage}
    \begin{minipage}{0.11\linewidth}
        \centering
        {\small \textbf{Target}}
        \textcolor{black}{\rule[1ex]{\linewidth}{2pt}}
        \includegraphics[width=1.0\textwidth]{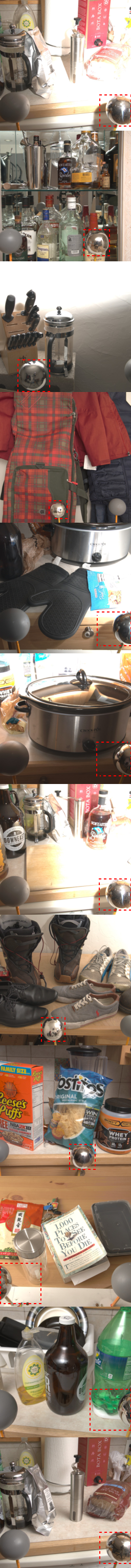}
    \end{minipage}
    \hfill
        \begin{minipage}{0.11\linewidth}
        \centering
        {\small \textbf{Input}}
        \textcolor{black}{\rule[1ex]{\linewidth}{2pt}}
        \includegraphics[width=1.0\textwidth]{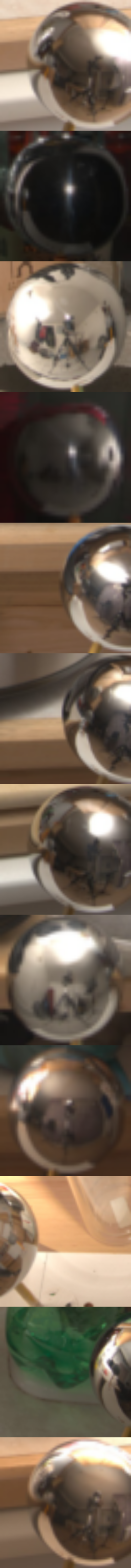}
    \end{minipage}
    \begin{minipage}{0.11\linewidth}
        \centering
        {\small \textbf{Ref}}
        \textcolor{black}{\rule[1ex]{\linewidth}{2pt}}
        \includegraphics[width=1.0\textwidth]{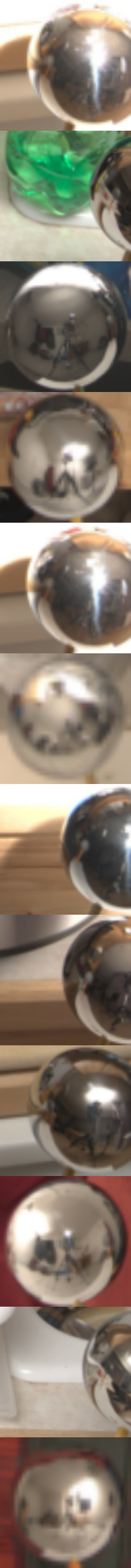}
    \end{minipage}
    \begin{minipage}{0.11\linewidth}
        \centering
        {\small \textbf{Ours}}
        \textcolor{black}{\rule[1ex]{\linewidth}{2pt}}
        \includegraphics[width=1.0\textwidth]{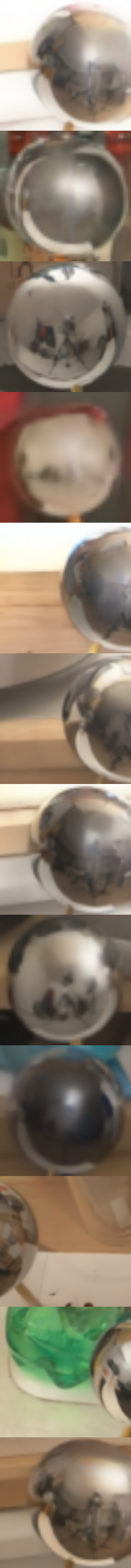}
    \end{minipage}
    \begin{minipage}{0.11\linewidth}
        \centering
        {\small \textbf{Target}}
        \textcolor{black}{\rule[1ex]{\linewidth}{2pt}}
        \includegraphics[width=1.0\textwidth]{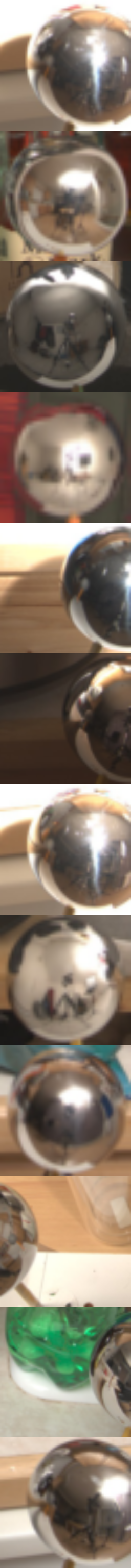}
    \end{minipage}
    \caption{We visualize more examples for the image relighting task in multi-illumination dataset\cite{murmann2019dataset}. \textbf{\textit{Right}}: Zoomed-in view of the chrome ball used as a probe to evaluate detail preservation in the environment map.}
    \label{fig:relight appendix}
\end{figure}
\newpage
\section*{NeurIPS Paper Checklist}

The checklist is designed to encourage best practices for responsible machine learning research, addressing issues of reproducibility, transparency, research ethics, and societal impact. Do not remove the checklist: {\bf The papers not including the checklist will be desk rejected.} The checklist should follow the references and follow the (optional) supplemental material.  The checklist does NOT count towards the page
limit. 

Please read the checklist guidelines carefully for information on how to answer these questions. For each question in the checklist:
\begin{itemize}
    \item You should answer \answerYes{}, \answerNo{}, or \answerNA{}.
    \item \answerNA{} means either that the question is Not Applicable for that particular paper or the relevant information is Not Available.
    \item Please provide a short (1–2 sentence) justification right after your answer (even for NA). 
\end{itemize}

{\bf The checklist answers are an integral part of your paper submission.} They are visible to the reviewers, area chairs, senior area chairs, and ethics reviewers. You will be asked to also include it (after eventual revisions) with the final version of your paper, and its final version will be published with the paper.

The reviewers of your paper will be asked to use the checklist as one of the factors in their evaluation. While "\answerYes{}" is generally preferable to "\answerNo{}", it is perfectly acceptable to answer "\answerNo{}" provided a proper justification is given (e.g., "error bars are not reported because it would be too computationally expensive" or "we were unable to find the license for the dataset we used"). In general, answering "\answerNo{}" or "\answerNA{}" is not grounds for rejection. While the questions are phrased in a binary way, we acknowledge that the true answer is often more nuanced, so please just use your best judgment and write a justification to elaborate. All supporting evidence can appear either in the main paper or the supplemental material, provided in appendix. If you answer \answerYes{} to a question, in the justification please point to the section(s) where related material for the question can be found.

IMPORTANT, please:
\begin{itemize}
    \item {\bf Delete this instruction block, but keep the section heading ``NeurIPS paper checklist"},
    \item  {\bf Keep the checklist subsection headings, questions/answers and guidelines below.}
    \item {\bf Do not modify the questions and only use the provided macros for your answers}.
\end{itemize}


\begin{enumerate}

\item {\bf Claims}
    \item[] Question: Do the main claims made in the abstract and introduction accurately reflect the paper's contributions and scope?
    \item[] Answer: \answerYes{} 
    \item[] Justification: We states our contributions in Section.~\ref{sec:intro}.

\item {\bf Limitations}
    \item[] Question: Does the paper discuss the limitations of the work performed by the authors?
    \item[] Answer: \answerYes{} 
    \item[] Justification: We discuss our limitations in Section.~\ref{sec:discuss}.

\item {\bf Theory Assumptions and Proofs}
    \item[] Question: For each theoretical result, does the paper provide the full set of assumptions and a complete (and correct) proof?
    \item[] Answer: \answerNA{} 
    \item[] Justification: Our paper does not include theoretical results. 

    \item {\bf Experimental Result Reproducibility}
    \item[] Question: Does the paper fully disclose all the information needed to reproduce the main experimental results of the paper to the extent that it affects the main claims and/or conclusions of the paper (regardless of whether the code and data are provided or not)?
    \item[] Answer: \answerYes{} 
    \item[] Justification: We provide experimental details in appendix section.~\ref{sec:exp_details}.

\item {\bf Open access to data and code}
    \item[] Question: Does the paper provide open access to the data and code, with sufficient instructions to faithfully reproduce the main experimental results, as described in supplemental material?
    \item[] Answer: \answerYes{} 
    \item[] Justification: We provide the link to our project page. 

\item {\bf Experimental Setting/Details}
    \item[] Question: Does the paper specify all the training and test details (e.g., data splits, hyperparameters, how they were chosen, type of optimizer, etc.) necessary to understand the results?
    \item[] Answer: \answerYes{} 
    \item[] Justification: We provide experimental details in appendix section ~\ref{sec:exp_details}.

\item {\bf Experiment Statistical Significance}
    \item[] Question: Does the paper report error bars suitably and correctly defined or other appropriate information about the statistical significance of the experiments?
    \item[] Answer: \answerNo{} 
    \item[] Justification: We report the averaged results after several runs.

\item {\bf Experiments Compute Resources}
    \item[] Question: For each experiment, does the paper provide sufficient information on the computer resources (type of compute workers, memory, time of execution) needed to reproduce the experiments?
    \item[] Answer: \answerYes{} 
    \item[] Justification: We provide those details in Section.~\ref{sec:exp_details}.
    
\item {\bf Code Of Ethics}
    \item[] Question: Does the research conducted in the paper conform, in every respect, with the NeurIPS Code of Ethics \url{https://neurips.cc/public/EthicsGuidelines}?
    \item[] Answer: \answerYes{} 
    \item[] Justification: Our research conform with the NeurIPS Code of Ethics.

\item {\bf Broader Impacts}
    \item[] Question: Does the paper discuss both potential positive societal impacts and negative societal impacts of the work performed?
    \item[] Answer: \answerNA{} 
    \item[] Justification: here is no societal impact of the work performed.
    
\item {\bf Safeguards}
    \item[] Question: Does the paper describe safeguards that have been put in place for responsible release of data or models that have a high risk for misuse (e.g., pretrained language models, image generators, or scraped datasets)?
    \item[] Answer: \answerNA{} 
    \item[] Justification: Our paper poses no such risks.

\item {\bf Licenses for existing assets}
    \item[] Question: Are the creators or original owners of assets (e.g., code, data, models), used in the paper, properly credited and are the license and terms of use explicitly mentioned and properly respected?
    \item[] Answer: \answerNA{} 
    \item[] Justification: Our paper does not use existing assets.

\item {\bf New Assets}
    \item[] Question: Are new assets introduced in the paper well documented and is the documentation provided alongside the assets?
    \item[] Answer: \answerNA{} 
    \item[] Justification: Our paper does not release new assets.

\item {\bf Crowdsourcing and Research with Human Subjects}
    \item[] Question: For crowdsourcing experiments and research with human subjects, does the paper include the full text of instructions given to participants and screenshots, if applicable, as well as details about compensation (if any)? 
    \item[] Answer: \answerNA{} 
    \item[] Justification: Our paper does not involve crowdsourcing nor research with human subjects.

\item {\bf Institutional Review Board (IRB) Approvals or Equivalent for Research with Human Subjects}
    \item[] Question: Does the paper describe potential risks incurred by study participants, whether such risks were disclosed to the subjects, and whether Institutional Review Board (IRB) approvals (or an equivalent approval/review based on the requirements of your country or institution) were obtained?
    \item[] Answer: \answerNA{} 
    \item[] Justification: Our paper does not involve crowdsourcing nor research with human subjects.

\end{enumerate}

\end{document}